\documentclass[final]{cvpr}

\usepackage{times}
\usepackage{epsfig}
\usepackage{graphicx}
\usepackage{amsmath}
\usepackage{amssymb}
\usepackage[binary-units=true]{siunitx}
\usepackage{caption}
\usepackage{subcaption}
\usepackage{booktabs}
\usepackage{multirow}
\usepackage{pstricks}
\usepackage{tabularx}
\usepackage{epstopdf}

\usepackage{amsmath,amsfonts,bm}

\def\eqref#1{equation~\ref{#1}}

\def\1{\bm{1}}

\def\rvd{{\mathbf{d}}}

\def\rvf{{\mathbf{f}}}

\def\rvh{{\mathbf{h}}}

\def\rvx{{\mathbf{x}}}

\DeclareMathAlphabet{\mathsfit}{\encodingdefault}{\sfdefault}{m}{sl}
\SetMathAlphabet{\mathsfit}{bold}{\encodingdefault}{\sfdefault}{bx}{n}

\def\sZ{{\mathbb{Z}}}

\newcommand{\R}{\mathbb{R}}

\DeclareMathOperator*{\argmax}{arg\,max}

\newcommand{\SKIP}[1]{}
\newcommand{\Skx}{S_\rvx^{(k)}}

\usepackage[pagebackref=true,breaklinks=true,colorlinks,bookmarks=false]
{hyperref}

\def\cvprPaperID{3221} 
\def\confYear{CVPR 2021}

\begin{document}

\title{Learning Optical Flow from a Few Matches}

\author{Shihao Jiang\textsuperscript{1,2,3} \qquad\qquad  Yao
Lu\textsuperscript{1,2,3}  \qquad\qquad Hongdong Li\textsuperscript{1,2}
\qquad\qquad Richard Hartley\textsuperscript{1,2}\\
\\
\textsuperscript{1}Australian National University \qquad \textsuperscript{2}ACRV
\qquad \textsuperscript{3}Data61, CSIRO
\\
{\tt\small \{shihao.jiang, yao.lu, hongdong.li, richard.hartley\}@anu.edu.au
}}

\maketitle

\begin{abstract}
\label{Sec:Abstract}
State-of-the-art neural network models for optical flow estimation require a dense correlation volume at high resolutions for representing per-pixel displacement. Although the dense correlation volume is informative for accurate estimation, its heavy computation and memory usage hinders the efficient training and deployment of the models. In this paper, we show that the dense correlation volume representation is redundant and accurate flow estimation can be achieved with only a fraction of elements in it. Based on this observation, we propose an alternative displacement representation, named Sparse Correlation Volume, which is constructed directly by computing the k closest matches in one feature map for each feature vector in the other feature map and stored in a sparse data structure. Experiments show that our method can reduce computational cost and memory use significantly, while maintaining high accuracy compared to previous approaches with dense correlation volumes. Code is available at \url{https://github.com/zacjiang/scv}. 
\end{abstract}

\section{Introduction}
\label{Sec:Intro}

\begin{figure}[t!]
     \centering
     \begin{subfigure}[b]{0.23\textwidth}
         \centering
         \includegraphics[width=\textwidth]{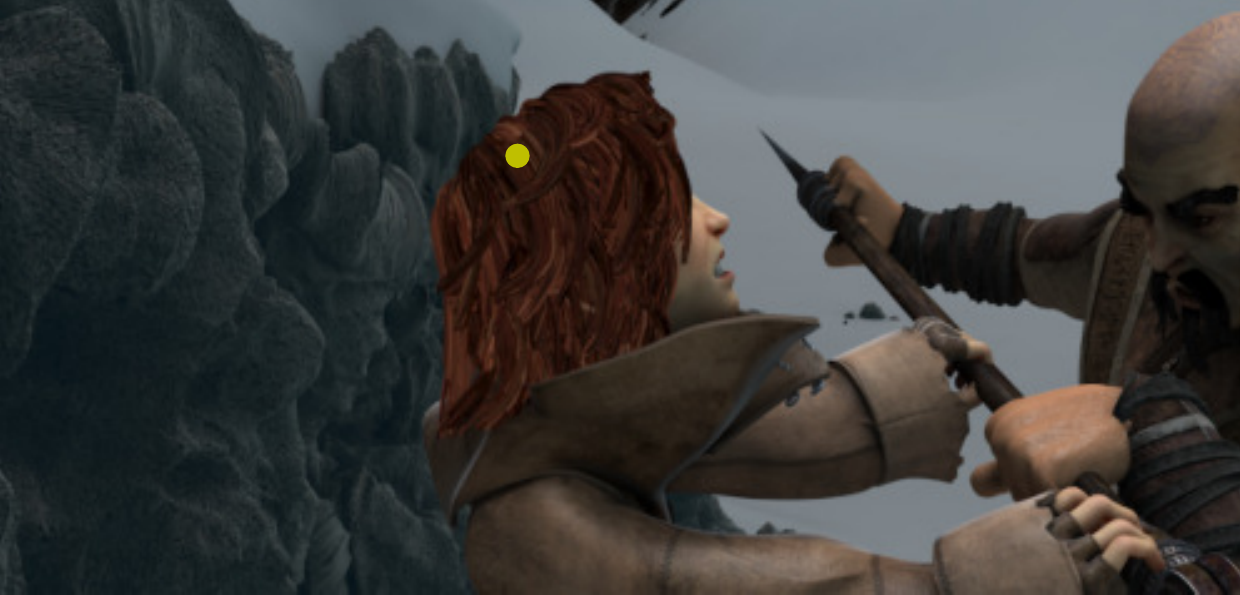}
         \caption{\em \small First image}
         \label{fig:input1}
     \end{subfigure}
     \hfill
     \begin{subfigure}[b]{0.23\textwidth}
         \centering
         \includegraphics[width=\textwidth]{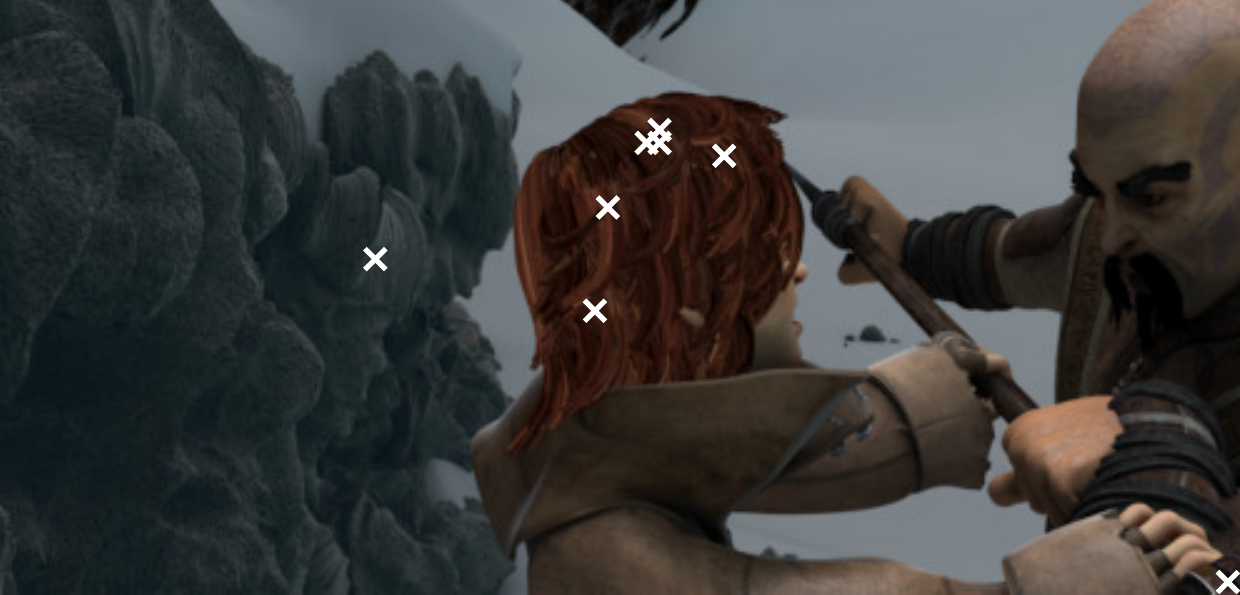}
         \caption{\em \small Second image}
         \label{fig:input2}
     \end{subfigure}

     \begin{subfigure}[b]{0.23\textwidth}
         \centering
         \includegraphics[width=\textwidth, trim=0 35 0 40, clip]{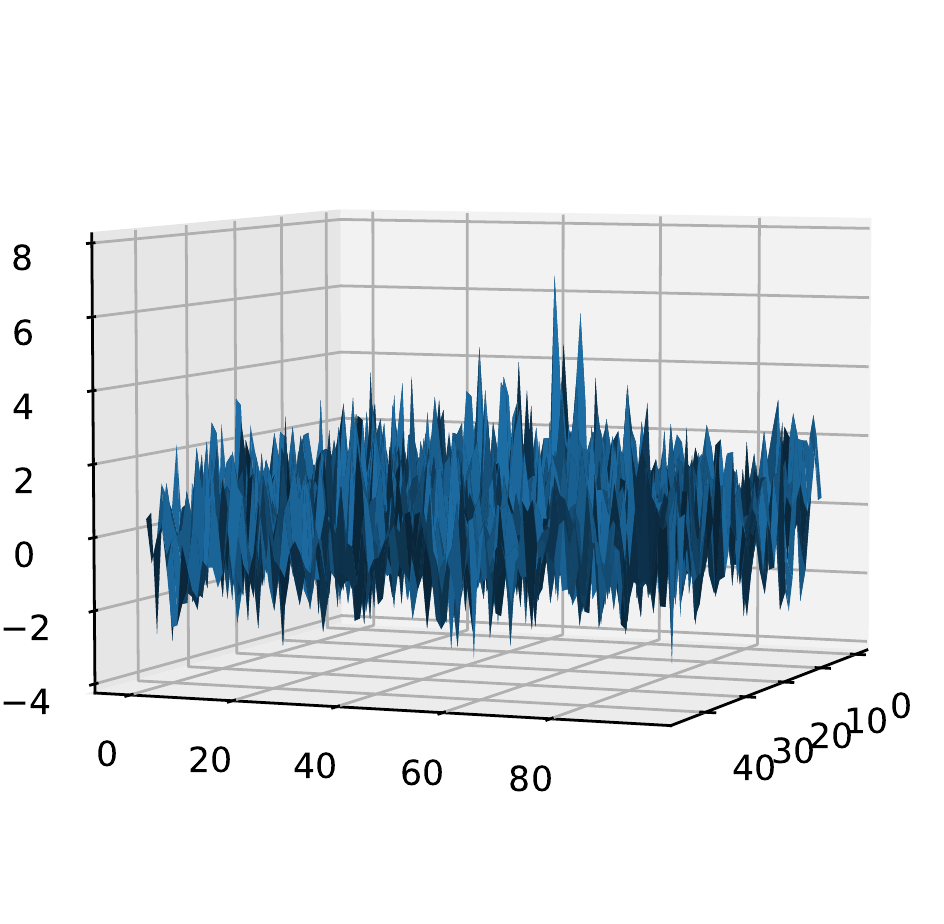}
         \caption{\em \small Dense correlation volume}
         \label{fig:dense3d}
     \end{subfigure}
     \hfill
     \begin{subfigure}[b]{0.23\textwidth}
         \centering
         \includegraphics[width=\textwidth, trim=0 35 0 40, clip]{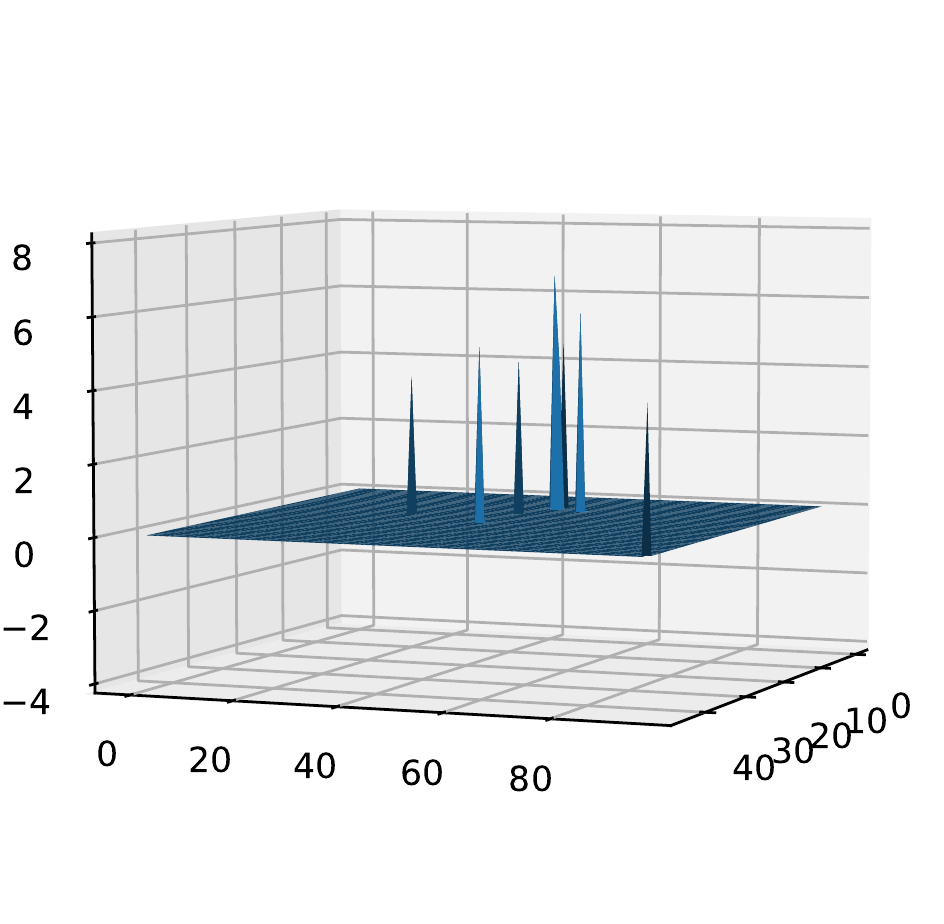} 
         \caption{\em \small Sparse correlation volume}
         \label{fig:sparse3d}
     \end{subfigure}
     
     \begin{subfigure}[b]{0.23\textwidth}
         \centering
         \includegraphics[width=\textwidth]{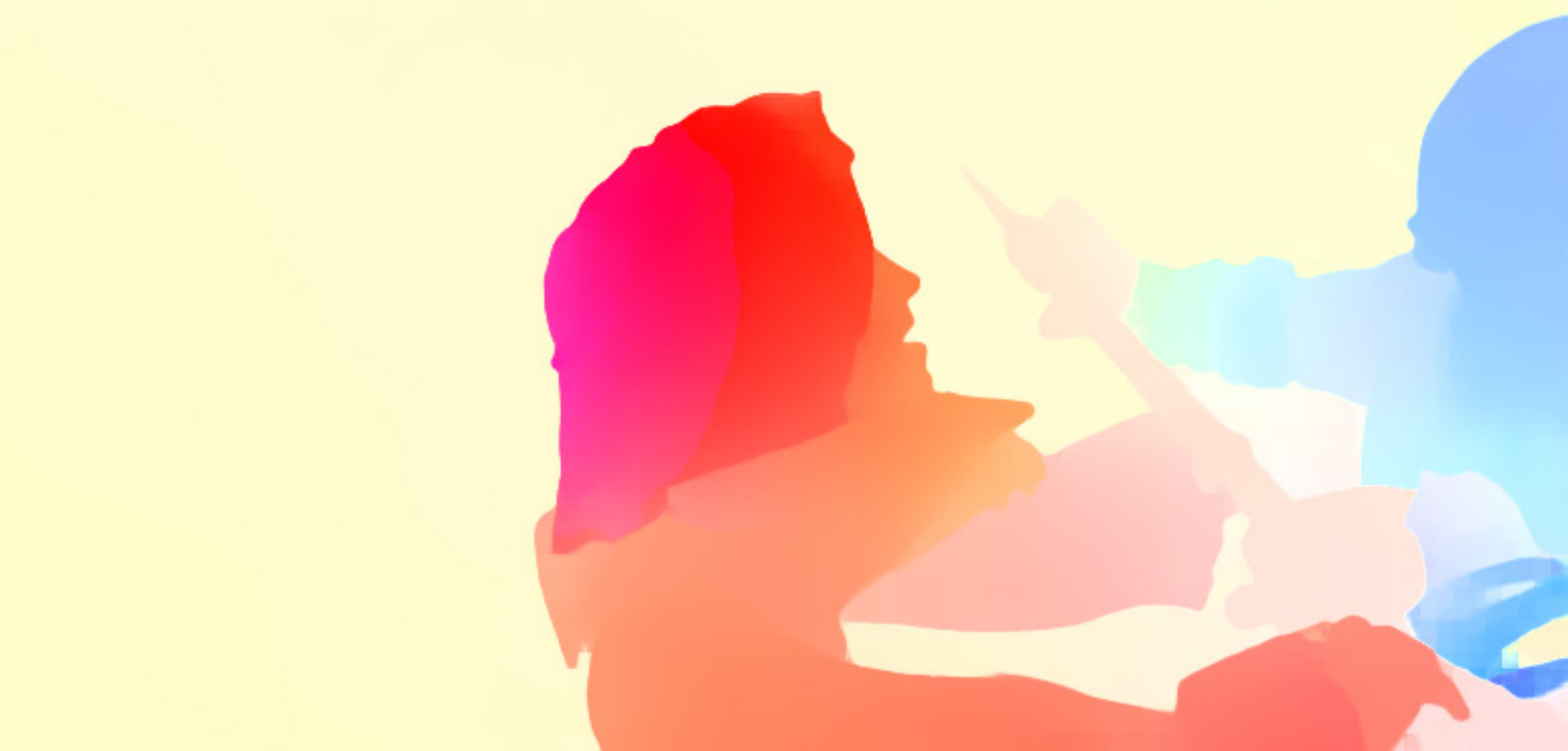}
         \caption{\em \small Optical flow (RAFT \cite{raft})}
         \label{fig:raft_result}
     \end{subfigure}
     \hfill
     \begin{subfigure}[b]{0.23\textwidth}
         \centering
         \includegraphics[width=\textwidth]{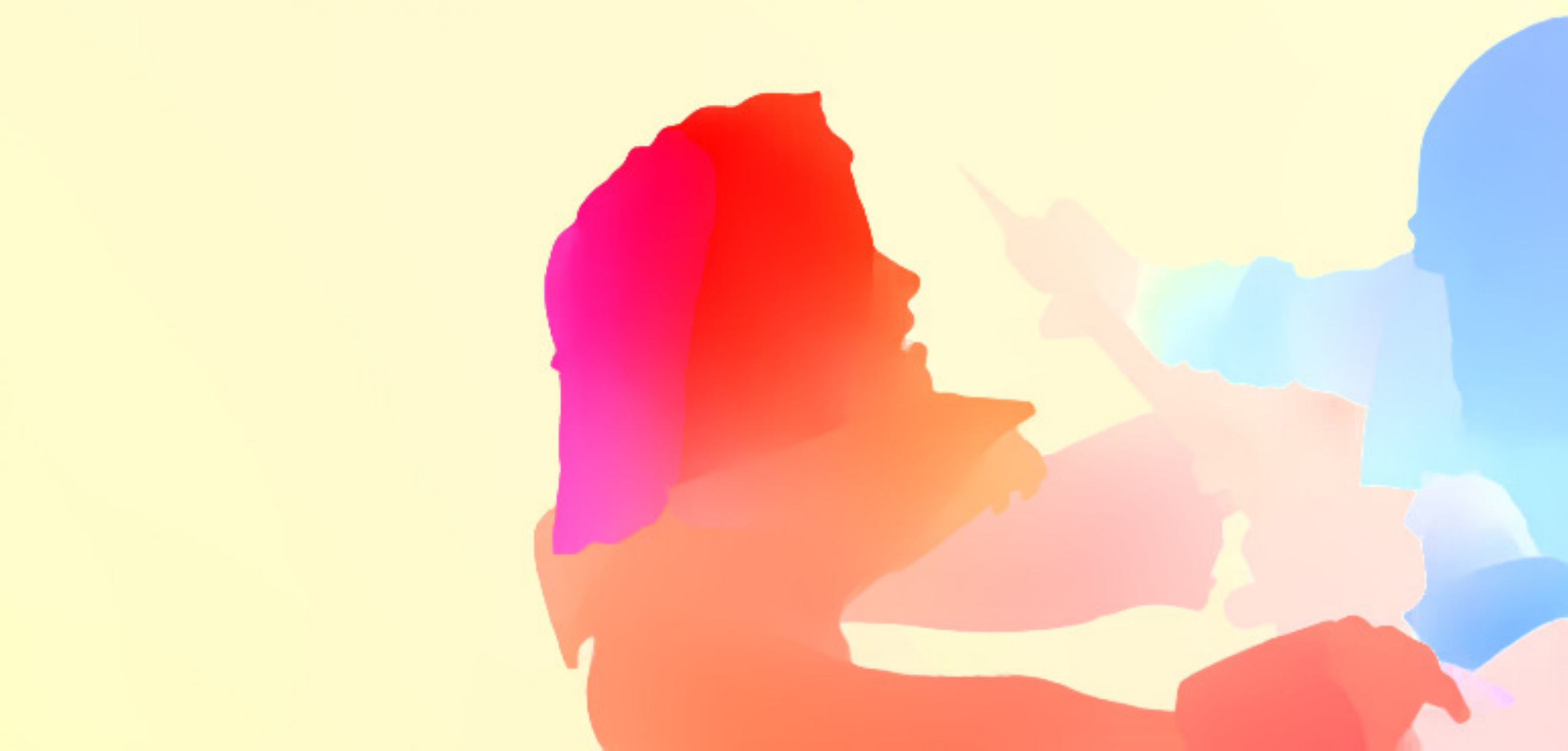} 
         \caption{\em \small Optical flow (ours)}
         \label{fig:ours_result}
     \end{subfigure}     
        \caption{ \textbf{Optical flow estimation with dense correlation volume and sparse correlation volume.} \em \small(c) and (d) illustrate the correlation volumes for a single pixel (yellow dot) in the first image. The white crosses in (b) indicate the top-$k$ matches. We show accurate optical flow can be estimated given only a few matching correlations.}
        \label{fig:demo}
\end{figure}

\begin{figure*}
     \centering
     \begin{subfigure}[b]{0.475\textwidth}
         \centering
         \includegraphics[width=\textwidth]{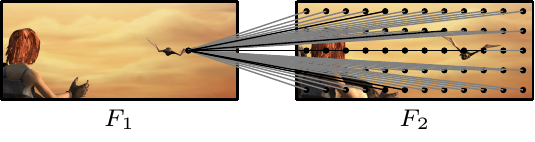}
         \caption{\em\small A dense correlation volume requires saving all pairs of matches. }
         \label{fig:dense}
     \end{subfigure}
    \hfill
    \begin{subfigure}[b]{0.475\textwidth}
         \centering
         \includegraphics[width=\textwidth]{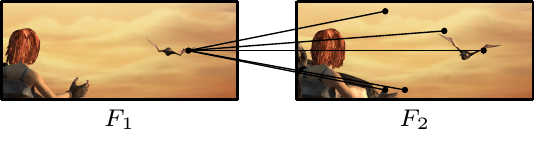}
         \caption{\em\small A sparse correlation volume requires saving only top-$k$ matches. }
         \label{fig:sparse}
     \end{subfigure}
    \caption{\textbf{Comparison between dense correlation volume and top-$k$ sparse correlation volume.} \em\small In a sparse correlation volume, only the top-$k$ matches are stored and the rest are discarded.}
    \label{fig:compare}
\end{figure*}

Optical flow estimation is a classic problem in computer vision \cite{horn}. 
It aims at finding pixelwise correspondences between two images. Traditionally it has been
formulated as an optimization problem solved by continuous 
\cite{brox, horn, xu2011motion} or discrete 
\cite{discreteflow, fullflow, dcflow} 
optimization. 
Since the development of deep learning, optical flow estimation has been formulated as a 
learning problem where direct regression from a neural network becomes a common approach 
\cite{flownet, flownet2}. 

One popular representation in dense correspondence problems is the correlation (cost)
volume, first introduced by Hosni \etal \cite{hosni2012fast}. 
Correlation volumes give an explicit representation of per-pixel displacements 
and have demonstrated their wide use in learning problems of stereo matching \cite{gcnet} 
and optical flow \cite{flownet, pwcnet}. 
Contrary to stereo matching problems, where the search space is along a scanline,
optical flow problems have a 2D search space, which leads to two challenges: 
large memory consumption and high computational cost when directly processing
a 4D volume. 

To reduce the memory and computational cost, existing approaches 
\cite{pwcnet, hd3, vcn, liteflownet, maskflownet} first build a feature pyramid and 
compute correlation volumes at
coarse resolutions, then gradually warp upper-level feature maps based on up-sampled flow
and construct a local correlation volume over a limited search range. 
One notable problem observed in previous work
\cite{brox,xu2011motion,devon}, 
was that coarse-to-fine frameworks fail to address the
case when the flow displacement is larger than the flow structure, i.e. the famous small 
objects moving fast problem. 

Recent approaches, Devon and RAFT \cite{devon, raft} proposed using direct search in 
the second image to remove the need for warping.
RAFT especially demonstrated the benefit of first constructing an all-pairs correlation 
volume and directly processing it at a single resolution rather than in a coarse-to-fine 
manner. However, the all-pairs correlation volume requires pair-wise dot product between
the two feature maps. Hence, both the time and the space complexity are $O(N^2)$,
where $N$ is the number of pixels of an image. A small $N$ is required to reduce the memory consumption
and therefore RAFT can only use $1/8$ resolution feature maps. 
A low-resolution feature map cannot fully represent the fine details of an image. We wonder
if there is a way to construct a correlation volume with the all-pairs search range 
but without exceeding the maximum GPU memory. We thus question the necessity of
storing all pairwise correlations and hypothesize that only storing the \emph{top-$k$}
correlations for each pixel might be sufficient. 

Our intuition is that a feature vector in one image has only a few  feature vectors in the other image with high correlation to match. Hence, there could be large redundancy
in the dense correlation volume where the small correlations do not contribute to the
prediction. 
Figure~\ref{fig:demo} illustrates the comparison between a dense correlation volume
and a sparse correlation volume. 

We propose a \emph{Sparse Correlation Volume} representation, where only the \emph{top-$k$}
correlations for each pixel are stored in a sparse data structure 
defined by a \{value, coordinates\} pair. 
In this paper, we demonstrate how a sparse correlation volume representation can
be used to solve the optical flow problem. We propose an approach to construct and 
process such a sparse correlation volume in an optical flow learning framework. 
We demonstrate that even if only a small fraction of elements are stored,
our results are still comparable to previous work \cite{raft} which employs a dense 
correlation volume. We finally demonstrate that the sparse approach allows the construction
of a high-resolution correlation volume, which can predict the motions of fine structures more
accurately than previous approaches. 

\begin{figure*}[ht!]
    \centering
    \includegraphics[width=\textwidth]{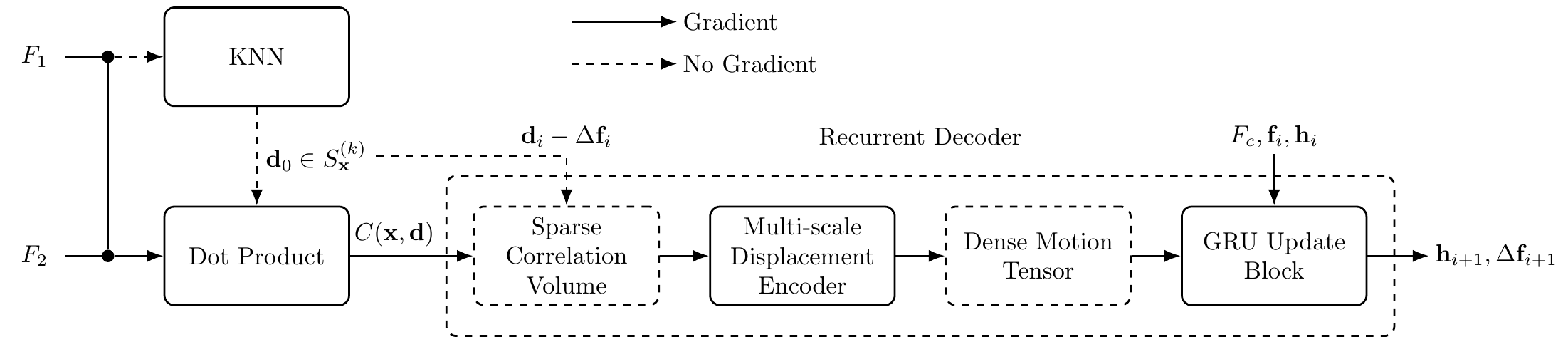}
    \caption{\textbf{Network architecture and residual flow prediction for a single iteration}. \em\small
    (1) $F_1$ and $F_2$ are feature maps extracted by a feature  extraction network. We form the 4D sparse correlation volume by first computing a set of displacements $\rvd_0 \in \widetilde{\mathcal{K}}$ with top-$k$ correlations with KNN. We then take the dot product for each feature vector in $F_1$ with its $k$ corresponding feature vectors in $F_2$. The dashed arrows denote paths that have no gradient flow while the solid arrows denote paths that do. 
    (2) In each iteration, the displacement vectors are updated by subtracting the residual flow $\rvd_i - \Delta \rvf_i$ to update the 4D correlation volume. A multi-scale displacement encoder is applied to encode the 4D sparse correlation volume to a 2D dense motion tensor. 
    (3) A GRU update block is applied to predict the residual flow $\Delta \rvf_{i+1}$ for the next iteration. The GRU block also takes input of $\rvh_i, F_c, \rvf_i$ which represents the hidden state vector of current iteration, feature map extracted by the context network and current estimation of optical flow, and outputs the hidden state vector for the next iteration as well the residual flow.} 
    \label{fig:architecture}
\end{figure*}

\section{Method}
\label{Sec:Method}

Let $I_1, I_2 : \sZ^2 \rightarrow \R^3$ be two RGB images. The problem is to
estimate a dense flow field $\rvf: \sZ^2 \rightarrow \R^2$ that maps each pixel
coordinate $\rvx$ to a displacement vector $\rvf (\rvx)$. 

In modern deep learning optical flow approaches, a feature extraction network is
first applied to extract feature maps from the image pair, $F_1, F_2 : \sZ^2
\rightarrow \R^c$, where $c$ is the number of channels. The correlation volume
$C : \sZ^4 \rightarrow \R$ is formed by computing inner products between
pairwise feature vectors, 
\begin{equation}
    C(\rvx, \rvd) = F_1(\rvx) \cdot F_2(\rvx + \rvd).
\end{equation}
The output is a four-dimensional tensor which can be represented as a set
\begin{equation}
    \mathcal{C} = \{C(\rvx, \rvd) \; | \; \rvx \in \mathcal{X}, \rvd \in
\mathcal{D} \}. 
\end{equation}
Here, ${\mathcal{X} = [0, h) \times [0, w) \cap \sZ^2}$ is the domain of the
feature map $F_1$ and ${|\mathcal{X}| = hw}$, where $h$ and $w$ represent the
height and width of $F_1$ respectively. The displacement set $\mathcal{D}$ is defined as
${\mathcal{D} = [-d, d]^2 \cap \sZ^2}$ where $d$ represents the maximum
displacement along the $x$ or $y$ direction and ${|\mathcal{D}| = (2d+1)^2}$.
Therefore, the correlation volume $\mathcal{C}$ contains ${hw(2d+1)^2}$
elements. 

To reduce the size of the correlation volume, previous approaches use
coarse-to-fine and warping methods to constrain the size of $d$ 
\cite{pwcnet,liteflownet,vcn}. To handle large displacements accurately, RAFT 
\cite{raft}
constructs an all-pairs correlation volume where the displacement range contains
the entire feature map. Excluding out-of-range matches, RAFT's all-pairs
correlation volume contains $N^2$ elements where $N = hw$. Therefore, lower-resolution 
feature maps are required to constrain $N$. In this work, we show that the
all-pairs correlation volume can in fact be a sparse tensor, where only a
small fraction of the values are stored and processed. We show that we can
effectively reduce the spatial complexity from $O(N^2)$ to $O(Nk)$ with only a minor drop of performance, 
where $k$ gives the number of matches we want to keep. The
main idea is demonstrated in Figure~\ref{fig:compare}.

\subsection{Sparse Correlation Volume}
\label{sec:scv}
For each $\rvx \in \mathcal{X}$, we define a set
\begin{equation}
     \Skx = \argmax_{S \subset \mathcal{D},
 |S| = k} \sum_{\rvd \in S} C(\rvx, \rvd)
\end{equation}
containing the $k$ displacements that give the maximum correlations. The
correlation volume can now be represented as a four-dimensional sparse tensor
\begin{equation}
    \widetilde{\mathcal{C}} = \{C(\rvx, \rvd) \; | \; \rvd \in
\Skx, \rvx \in \mathcal{X} \}.
\end{equation}
This sparse correlation volume contains ${hwk}$ elements as opposed to the
original dense correlation volume with ${h^2w^2}$ elements. The constant $k$ is
typically a small number (e.g. $k=8$). 

We now show how to construct the sparse correlation volume and
estimate optical flow from it. Our network architecture is shown in
Figure~\ref{fig:architecture}. 

\subsection{$k$-Nearest Neighbours}
\label{sec:knn}
We first use two weight-sharing feature extraction networks to extract $1/4$
resolution feature maps from the input images. 
Our feature extraction networks consist of six residual blocks and the number of feature
channels is $256$.
To construct the sparse correlation volume, we use a $k$-nearest neighbours ($k$NN) module \cite{faiss} to
compute a set of indices with the $k$ largest correlation scores for each feature vector
in $F_1$.
The sparse correlation volume is computed by taking the dot product between each
feature vector in $F_1$ with the top $k$ feature vectors given by the indices in
$F_2$. During back-propagation, the gradients are only propagated to the $k$
feature vectors that are selected by the $k$NN module.

\begin{figure}[ht!]
    \begin{subfigure}[b]{0.475\textwidth}
    \centering
    \includegraphics[width=\textwidth]{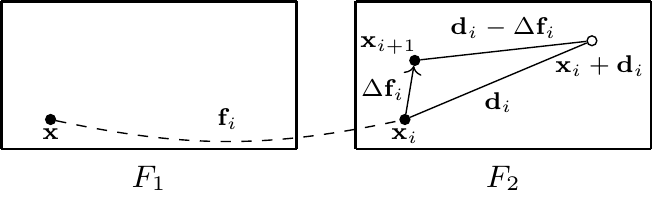}
    \vspace{-0.6cm}
    \caption{\em\small Displacements update. As a pixel's coordinates are updated by adding $\Delta \mathbf{f}_i$, the relative displacements are diminished by $\Delta \mathbf{f}_i$.}
    \vspace{0.7cm}
    \label{fig:disp_update}
    \end{subfigure}
    \hfill 
    \begin{subfigure}[b]{0.475\textwidth}
    \centering
    \includegraphics[width=0.8\textwidth]{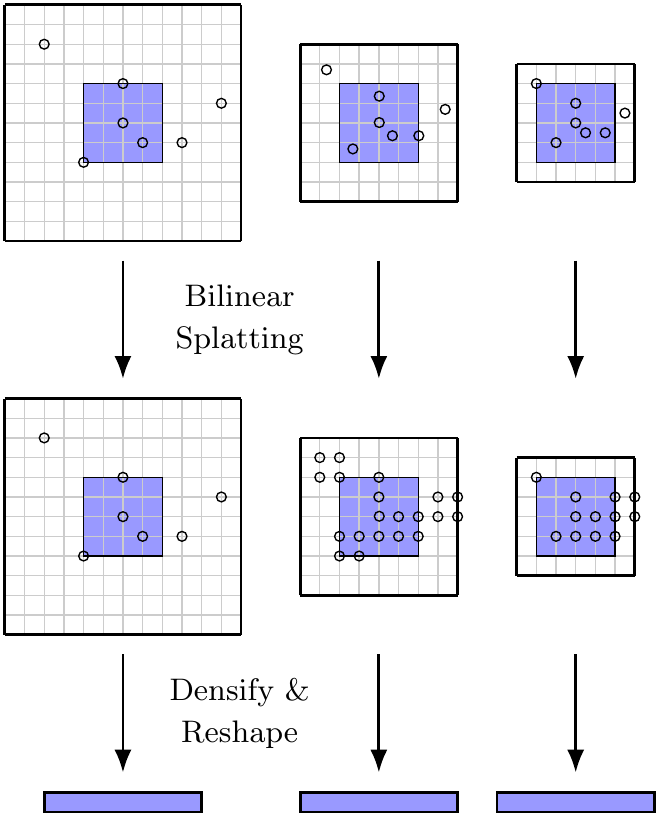}
    \caption{\em \small Multi-scale displacement encoder. We first form a multi-level sparse correlation pyramid by scaling the coordinates by different constants. We then bilinearly splat the correlations onto the integer grids and extract correlation values within a local window. The extracted windows are converted to dense tensors and are reshaped and concatenated to form a single $h \times w \times c$ tensor. }
    \label{fig:encoder}
    \end{subfigure}
    \caption{\textbf{Illustration of how to process a sparse correlation volume in an iterative fashion.}}
    

\end{figure}
\subsection{Displacements Update}
\label{sec:disp_update}
We adopt an overall iterative residual refinement approach. As shown by previous
work \cite{irr, raft}, estimating residual flows can effectively reduce the
search space and predict better results than direct regression 
\cite{flownet2,flownet}. Rather than directly predicting optical flow $\rvf$, a residual flow
$\Delta \rvf_{i+1}$ is predicted at each step and used to update the current flow
estimation $\rvf_{i+1} = \rvf_i + \Delta \rvf_{i+1}$. 

At each step, a pixel $\rvx$ in $F_1$ is mapped to $\rvx_i$ in $F_2$ according
to the current estimate of flow $\rvx_i = \rvx + \rvf_i$.
Our sparse correlation volume $\widetilde{\mathcal{C}}$ described in Section~\ref{sec:scv}
can be regarded as an initial estimation $\rvf_0 = 0$ at $i=0$.
When the coordinate $\rvx_i$ is updated to $\rvx_{i+1} = \rvx_i + \Delta
\rvf_i$, the relative displacements in $\widetilde{\mathcal{C}}$ should be updated
accordingly as well.
To do so, we shift the coordinates of the sparse correlation tensor by
subtracting $k$-nearest neighbouring $\Delta \rvf_i$ from $\rvd_i$ in each step, $C_i(\rvx, \rvd_i) =
C_{i+1}(\rvx, \rvd_i - \Delta \rvf_i)$, as depicted in
Figure~\ref{fig:disp_update}. 
Note here we allow $\rvd_i - \Delta \rvf_i$ to be floating-point.
It is also important to note that the inner products are computed only once at
the start since in each step, only the correlation coordinates change while the
correlation values remain the same.

\subsection{Multi-scale Displacement Encoder}
One question that is often raised with any sparse approach is how to process a
sparse tensor, since the regularity of a normal dense $h \times w \times c$
tensor is lost. Sparse convolutions \cite{minkowski} may be used, however, we
will present a simpler and more efficient approach here. 

A dense all-pairs correlation volume has dimension $h \times w \times h
\times w$ and we have reduced it to a sparse tensor with $h \times w \times k$
elements where only the top-$k$ correlations for each pixel are saved. We can
see that the first two dimensions are still dense and what has become sparse
are the third and fourth dimensions. The goal here is to encode the $k$ elements
for each pixel and form a dense $h \times w \times c$ tensor, which can later be
used to predict $\Delta \rvf_{i+1}$. 

Following previous work \cite{raft}, we propose creating multi-scale sparse
tensors and sampling displacements locally with a fixed radius at different
resolutions. 
Coarser resolutions give larger context while finer resolutions give more
accurate displacements.
We then convert the sparse tensors at each level to dense tensors and
concatenate them to form a single 2D tensor. 
This is illustrated in Figure~\ref{fig:encoder}.

At each iteration $i$, for each pixel $\rvx$, we start with a set of the top $k$ correlation
positions $\Skx$.  So, the set 
$\{\big(\rvd, C(\rvx, \rvd)\big) ~|~ \rvd \in \Skx \}$ records
the top $k$ correlation values for pixel $\rvx$ and their locations,
obtained using a $k$NN algorithm.

We construct a five-level sparse correlation volume pyramid by dividing the coordinates by $(1, 2, 4, 8, 16)$ and denote the scaled displacements at level $l$,
updated with the current $\Delta \rvf_i$ by
$\rvd^l = \left(\rvd_i - \Delta \rvf_i\right) ~/~ 2^{l-1}$.

In addition, we denote the correlation values at level $l$ by
$C^l(\rvx, \rvd^l) = C(\rvx, \rvd)$ for $\rvd \in \Skx$.
At each level, we constrain the displacements by a constant radius $r$
and define the windowed set of correlation values at level $l$,
\begin{equation}
\label{eq:level-l-correlations}
    \big\{\big(\rvd^l, C^l(\rvx, \rvd^l)\big) 
    ~\big |~ \|\rvd^l\|_\infty \le r,\, \rvd \in \Skx  \big\} ~.
\end{equation}
\SKIP{
\begin{equation}
    \mathcal{D}^l_r = \left\{ \rvd^l \; | \; \|\rvd^l||_\infty \leq r \right\},
\end{equation}
where the maximum $x$ or $y$-direction displacement is less than $r$. 
We thus have a sparse correlation volume pyramid
$\left\{\mathcal{C}^l\right\}_{l = 1 \dots 5}$ where
\begin{equation}
    \mathcal{C}^l = \left\{C(\rvx, \rvd) \; | \; \rvd \in \mathcal{D}^l_r,~ \rvx
\in \mathcal{X} \right\} ~.
\end{equation}
} 
Since the coordinates $\rvd^l$ are not necessarily integers, we need to resample to integer coordinates in order to densify the sparse tensor 
of correlations. 
We propose an approach which we call ``bilinear splatting". The correlation
values are bilinearly splatted to the four nearest integer grids. 
For instance, the correlation $C^l(\rvx, \rvd^l)$ at location $\rvd^l$
is propagated to each of four neighbouring integer points, denoted
by $[\rvd^l] = (d_x, d_y)$, according to
\begin{equation*}
    C^l(\rvx, [\rvd^l]) = \left(1-|d_x^l -  d_x|
\right) \left( 1 - |d_y^l - d_y| \right)C^l(\rvx, \rvd^l).
\end{equation*}
These values are then summed for the set
of correlations (\ref{eq:level-l-correlations})
and the sparse tensors of each level are
converted to dense tensors, reshaped and concatenated to form a 
single 2D dense tensor of dimension $5 (2r+1)^2$, where $5$
is the number of pyramid levels.

The approach we introduce here does not require learning. It is merely a
conversion between sparse and dense tensors hence is simpler than sparse
convolutions. 

\subsection{GRU Update Block}
Each vector in this 2D dense tensor encodes position information as well as the
correlation values of the $k$ matches. We concatenate the 2D motion tensor with
the context features and current estimate of flow and pass it through a gated recurrent units (GRU)
update block. The GRU update block estimates the residual flow $\Delta
\rvf_{i+1}$ which is used to shift the correlation volume coordinates in the
next step. 

\setlength\tabcolsep{.7em}
\begin{table*}[ht!]
\centering
\resizebox{\textwidth}{!}{
\begin{tabular}{clccccccc}
\toprule
\multirow{3}{*}{Training Data} & \multirow{3}{*}{Method} & \multicolumn{2}{c}{Sintel (train)} &  \multicolumn{2}{c}{KITTI-15 (train)} & \multicolumn{2}{c}{Sintel (test)} & \multicolumn{1}{c}{KITTI-15 (test)} \\
\cmidrule(lr){3-4}
\cmidrule(lr){5-6}
\cmidrule(lr){7-8}
\cmidrule(lr){9-9}
& & Clean & Final & EPE & F1-all & Clean & Final & F1-all \\

\midrule    
\multirow{9}{*}{C + T} 
                       & LiteFlowNet2\cite{liteflownet2}   & 2.24  & 3.78  & 8.97 & 25.9 & - & - & - \\
                       & VCN\cite{vcn}            & 2.21  & 3.68  & 8.36 & 25.1 & - & -     & - \\ 
                       & MaskFlowNet\cite{maskflownet} & 2.25 & 3.61 & - & 23.1 & - & - & - \\ 
                       & FlowNet2\cite{flownet2}       & 2.02  & 3.54 & 10.08 & 30.0 & 3.96  & 6.02 & - \\
                       & DICL\cite{dicl}        & 1.94 & 3.77 & 8.70 & 23.6 & - & - & - \\
                       & RAFT\cite{raft}        & 1.43 & \textbf{2.71} & \textbf{5.04} & \textbf{17.4} & - & - & - \\ 
                       & Ours & \textbf{1.29} & 2.95 & 6.80 & 19.3 & - & - & - \\
                       \midrule
\multirow{13}{*}{C+T+S/K(+H)} 
                     & FlowNet2 \cite{flownet2}  & (1.45) & (2.01) & (2.30) & (6.8) & 4.16  & 5.74 & 11.48  \\
                     & PWC-Net+\cite{pwcnet+}   & (1.71)     & (2.34)  & (1.50) & (5.3)  & 3.45  & 4.60 & 7.72 \\
                     & LiteFlowNet2 \cite{liteflownet2} & (1.30) & (1.62) & (1.47) & (4.8) & 3.48  & 4.69 & 7.74 \\
                     & HD3 \cite{hd3}         & (1.87)     & (1.17) & (1.31) & (4.1)  & 4.79  & 4.67 & 6.55 \\
                     & IRR-PWC \cite{irr}     & (1.92) & (2.51) & (1.63) & (5.3) & 3.84  & 4.58  & 7.65 \\
                     & VCN \cite{vcn}            & (1.66)     & (2.24) & (1.16) & (4.1) & 2.81  & 4.40 & 6.30 \\
                     & MaskFlowNet\cite{maskflownet} & - & - & - & - & 2.52 & 4.17 & 6.10 \\
                     & ScopeFlow\cite{scopeflow} & - & - & - & - & 3.59 & 4.10 & 6.82 \\
                     & DICL\cite{dicl}  & (1.11) & (1.60) & (1.02) & (3.6) & 2.12 & 3.44  & 6.31 \\
                     & RAFT\cite{raft} (warm-start) & (0.77) & (1.27) & - & - & \textbf{1.61} & \textbf{2.86}  & - \\
                     & RAFT\cite{raft} (2-view) & (0.76)  & (1.22) & (0.63) & (1.5) & 1.94 & 3.18 & \textbf{5.10} \\ 
                     & Ours (warm-start) & (0.86) & (1.75) & - & - & 1.77 & 3.88 & - \\
                     & Ours (2-view) & (0.79) & (1.70) & (0.75) & (2.1) & 1.72 & 3.60 & 6.17 \\
                     \bottomrule
\end{tabular}
}
\caption{\textbf{Quantitative results on Sintel and KITTI 2015 datasets.} \em\small EPE refers to the average endpoint error and F1-all refers to the percentage of optical flow outliers over all pixels. ``C + T'' refers to results that are pre-trained on Chairs and Things datasets. ``S/K(+H)'' refers to methods that are fine-tuned on Sintel, KITTI and some on HD1K datasets. Paratheses refer to the training results and the best results are in bold font. 
}
\label{Tab:Results}
\end{table*}

\begin{figure*}[ht!]
     \centering
     
     \begin{subfigure}[b]{0.245\textwidth}
         \centering
         \caption*{Input image 1}
         \includegraphics[width=\textwidth]{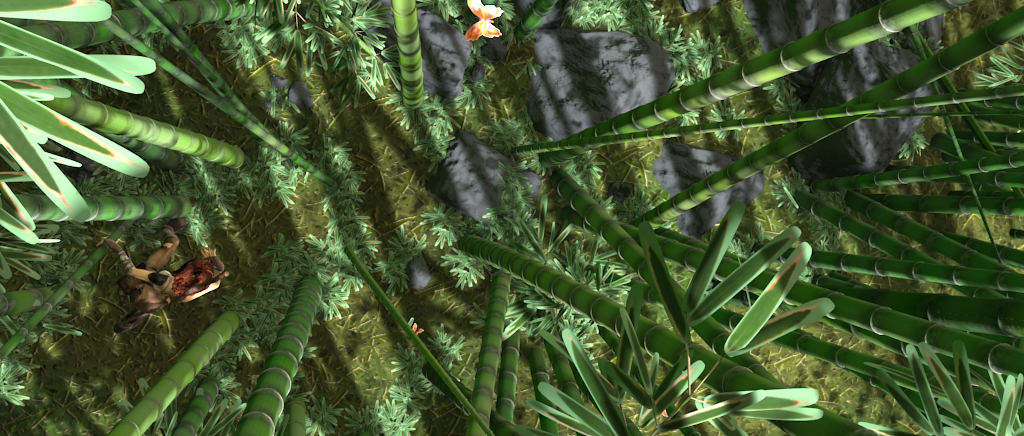}
     \end{subfigure}
     \hfill
     \begin{subfigure}[b]{0.245\textwidth}
         \centering
         \caption*{Ground-truth}
         \includegraphics[width=\textwidth]{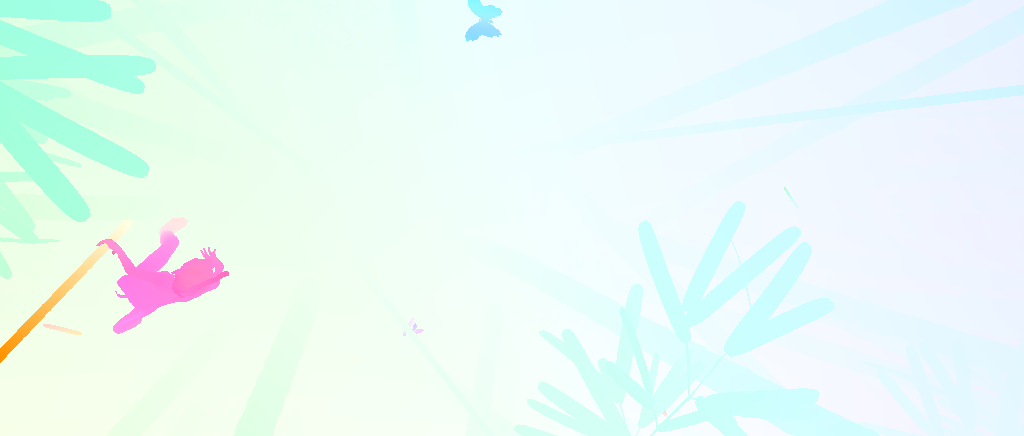}
     \end{subfigure}
     \hfill
     \begin{subfigure}[b]{0.245\textwidth}
         \centering
         \caption*{RAFT \cite{raft}}
         \includegraphics[width=\textwidth]{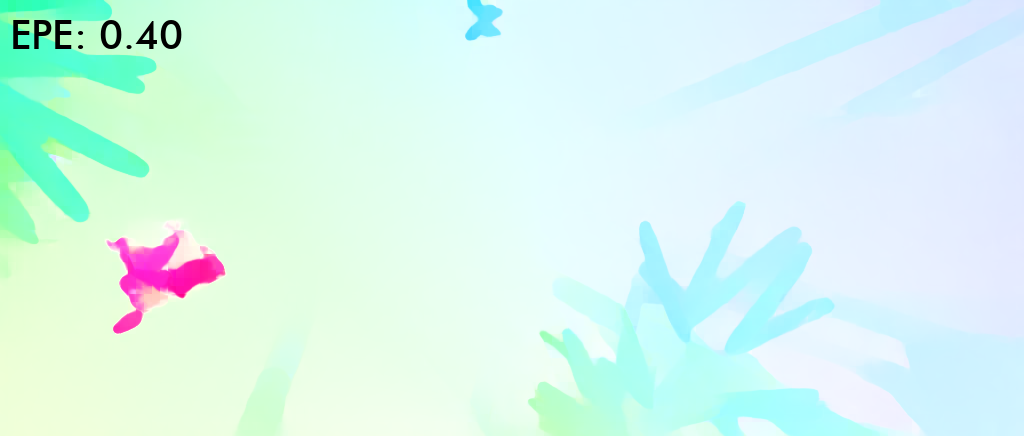}
     \end{subfigure}
     \hfill
     \begin{subfigure}[b]{0.245\textwidth}
         \centering
         \caption*{Ours}
         \includegraphics[width=\textwidth]{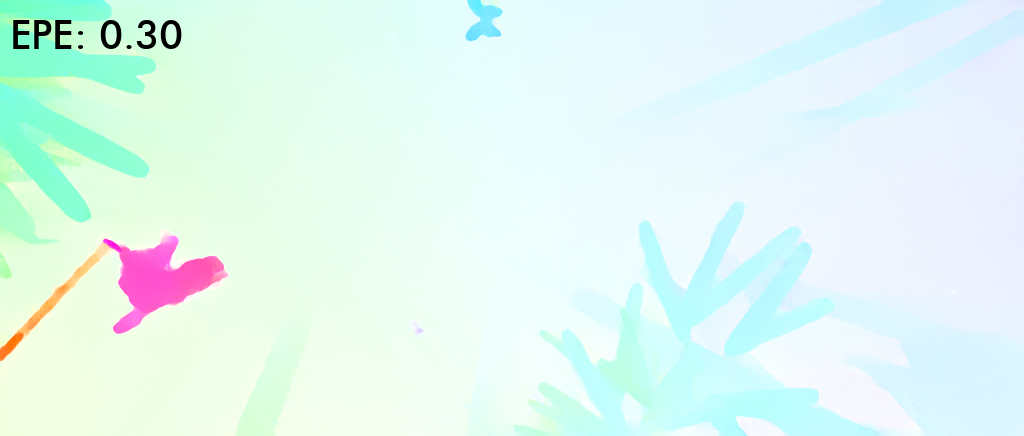}
     \end{subfigure}

     \begin{subfigure}[b]{0.245\textwidth}
         \centering
         \includegraphics[width=\textwidth]{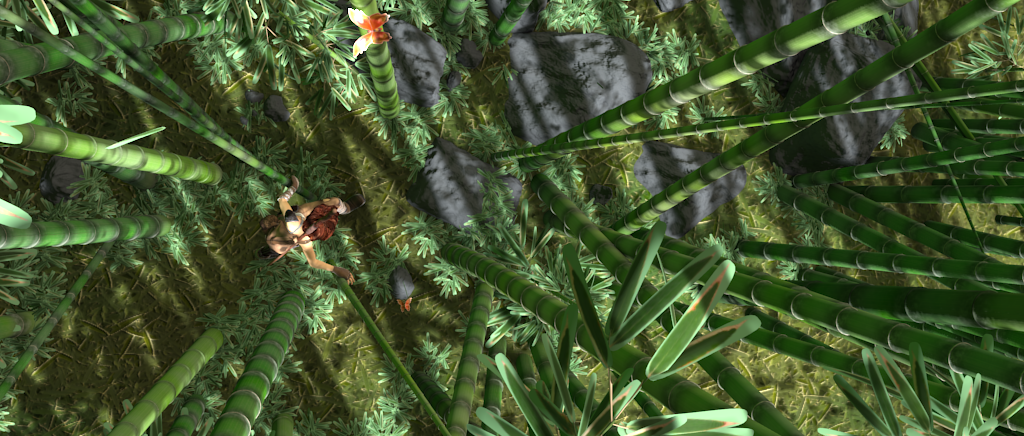}
     \end{subfigure}
     \hfill
     \begin{subfigure}[b]{0.245\textwidth}
         \centering
         \includegraphics[width=\textwidth]{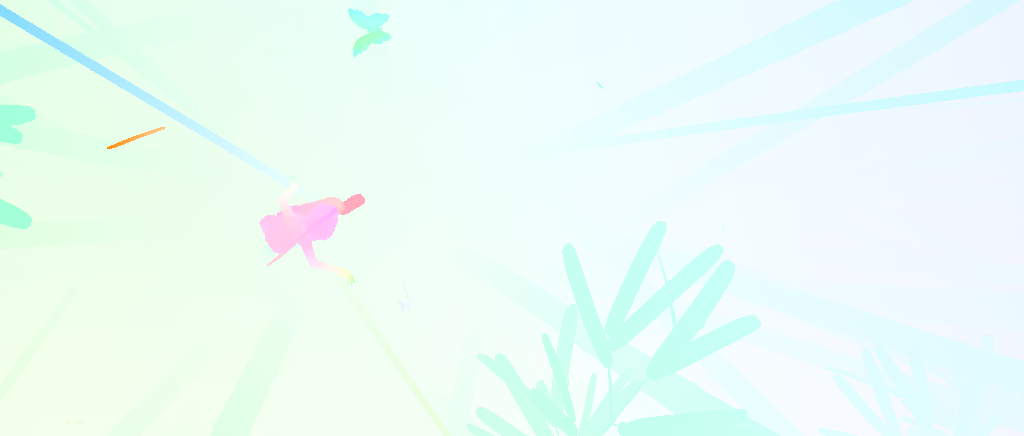}
     \end{subfigure}
     \hfill
     \begin{subfigure}[b]{0.245\textwidth}
         \centering
         \includegraphics[width=\textwidth]{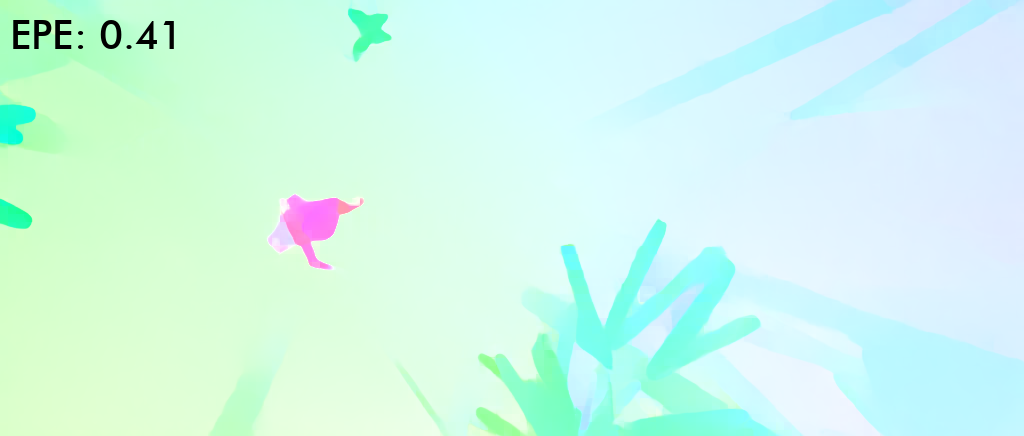}
     \end{subfigure}
     \hfill
     \begin{subfigure}[b]{0.245\textwidth}
         \centering
         \includegraphics[width=\textwidth]{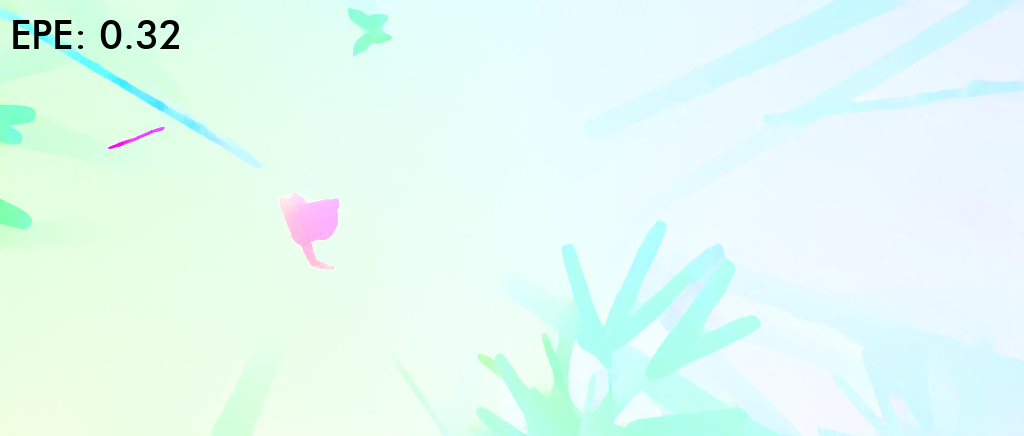}
     \end{subfigure}
     
     \begin{subfigure}[b]{0.245\textwidth}
         \centering
         \includegraphics[width=\textwidth]{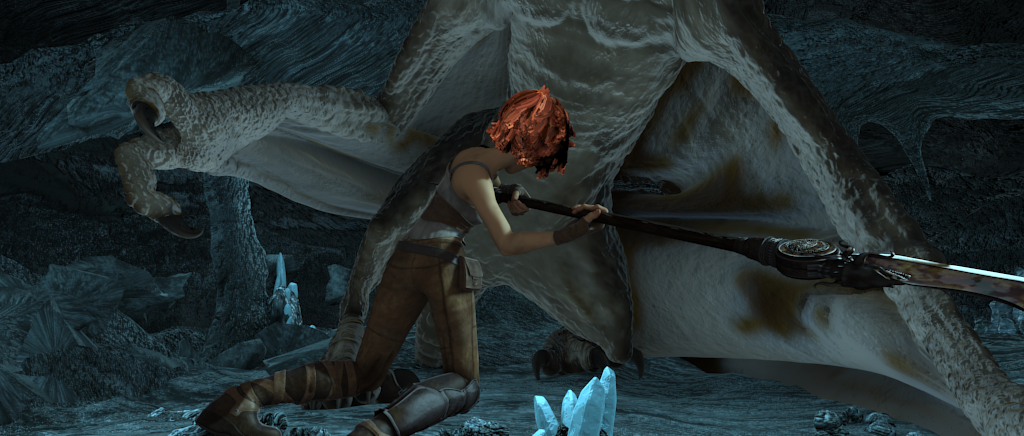}
     \end{subfigure}
     \hfill
     \begin{subfigure}[b]{0.245\textwidth}
         \centering
         \includegraphics[width=\textwidth]{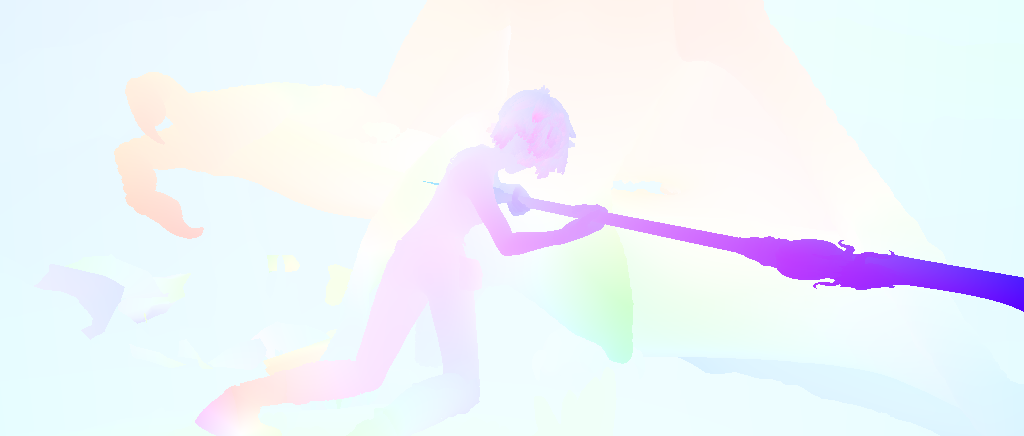}
     \end{subfigure}
     \hfill
     \begin{subfigure}[b]{0.245\textwidth}
         \centering
         \includegraphics[width=\textwidth]{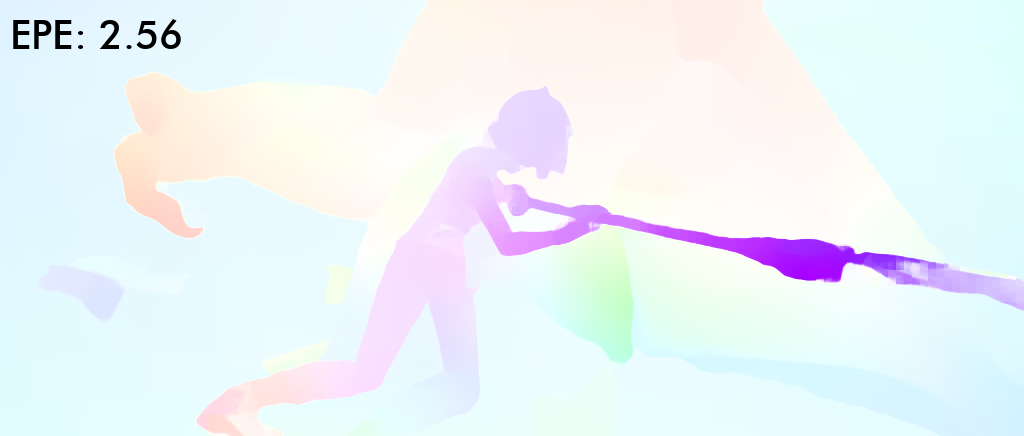}
     \end{subfigure}
     \hfill
     \begin{subfigure}[b]{0.245\textwidth}
         \centering
         \includegraphics[width=\textwidth]{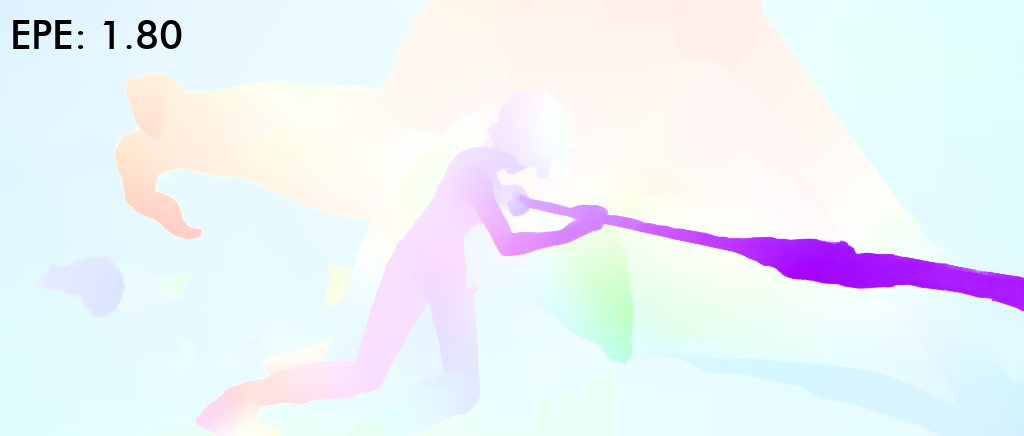}
     \end{subfigure}
    
     \begin{subfigure}[b]{0.245\textwidth}
         \centering
         \includegraphics[width=\textwidth]{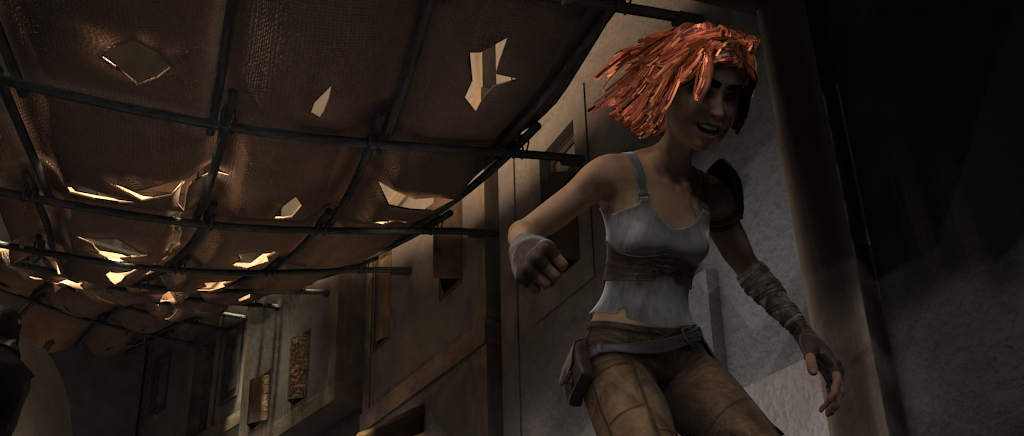}
     \end{subfigure}
     \hfill
     \begin{subfigure}[b]{0.245\textwidth}
         \centering
         \includegraphics[width=\textwidth]{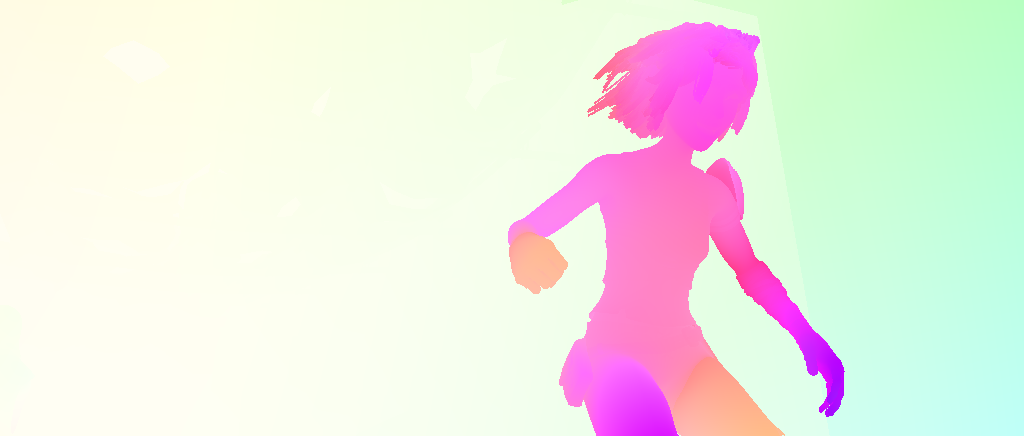}
     \end{subfigure}
     \hfill
     \begin{subfigure}[b]{0.245\textwidth}
         \centering
         \includegraphics[width=\textwidth]{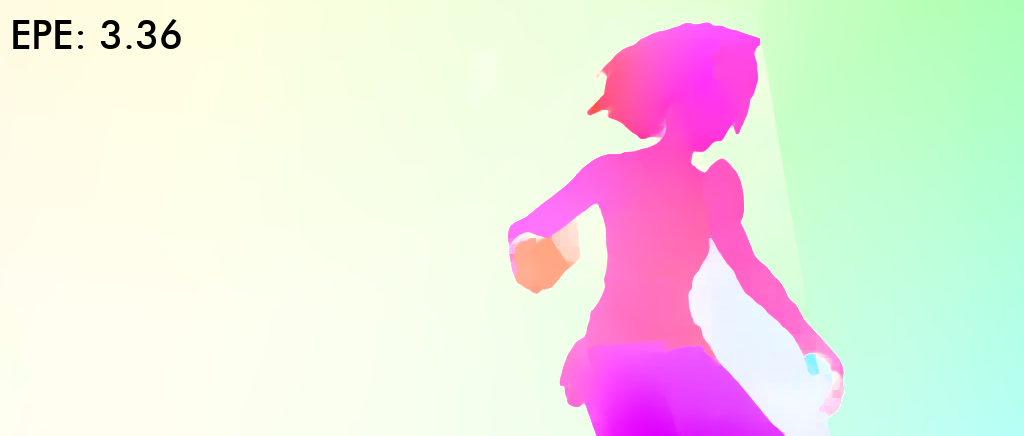}
     \end{subfigure}
     \hfill
     \begin{subfigure}[b]{0.245\textwidth}
         \centering
         \includegraphics[width=\textwidth]{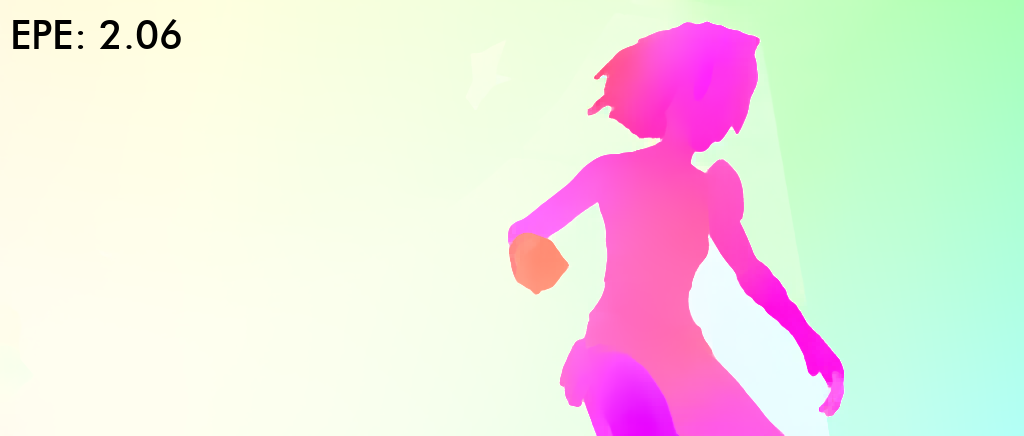}
     \end{subfigure}
     
    \caption{\textbf{Qualitative results on Sintel.} \em\small We compare the results of the pre-training models (trained
    on Chairs + Things) on the Sintel training dataset. The results are 
    compared against RAFT. We demonstrate 
    cases where the quarter resolution correlation volume outperforms the
    eighth resolution correlation volume. Two noticable examples are the 
    first two rows, where the motion of a thin bamboo cannot be captured 
    by the eighth resolution correlation volume due to the large 
    downsampling. Nevertheless, it can be accurately predicted by our 
    method. Best viewed on screen when zooming in.}
    \label{fig:vis}
\end{figure*}

\section{Experiments}
\label{Sec:Experiments}

\subsection{Implementation details}
\label{Sec:Implementaion}
\paragraph{Network details} We first extract quarter-resolution feature maps
with $256$ channels. Our feature extraction network contains six residual
blocks. 
When passing the feature maps to the $k$NN, we set $k=8$. Namely, for each feature vector,
return the indices of the the top-$8$ feature vectors that give the maximum inner products.
The GRU update block takes the current estimate of optical flow as well as
the context feature map as input. The context feature map is extracted by a
separate network with $128$ channels. The GRU update block also updates a
128-dimensional hidden-state feature vector. 
During training time, the GRU iterates $8$ times as opposed to $12$ times in RAFT \cite{raft}. 

\paragraph{Training schedule}
Following previous work, we first pre-train our model on FlyingChairs
\cite{flownet} for 120k iterations with batch size 6 and then on FlyingThings
\cite{things} for another 120k iterations with batch size 4. We then fine-tune on
a combination of Things, Sintel \cite{sintel}, KITTI 2015 \cite{kitti} and HD1K \cite{hd1k} 
for 120k iterations for Sintel evaluation and 50k on KITTI 2015 \cite{kitti} for KITTI evaluation. 
We use a batch size of 4 for fine-tuning. We train our model on two 2080Ti GPUs. The ablation
experiments are conducted on a single Tesla P100 GPU. We implemented with the 
\texttt{PyTorch} library \cite{pytorch}. 

\paragraph{Loss function}
Similar to RAFT \cite{raft}, we employ a recurrent network architecture where a sequence of 
residual flows $\Delta \rvf_i$ are predicted. The optical flow prediction in each step can be
represented as $\rvf_{i+1} = \rvf_i + \Delta \rvf_{i+1}$ and the initial values are
$\rvf_0 = 0,~\Delta \rvf_0 = 0$.

We apply the loss function on the sequence of optical flow predictions. 
Given the ground-truth optical flow $\rvf_{\text{gt}}$ and predicted optical flow at each step
$\rvf_i$, the loss function is defined as 
%
\begin{equation*}
    L = \sum_{i = 1}^N \gamma^{N - i} \|\rvf_i - \rvf_{\text{gt}} \|_1.
\end{equation*}
The weight $\gamma$ is set to 0.8 for pre-training on Chairs and Things and 0.85 for fine-tuning 
on Sintel and KITTI. The total number of steps $N$ is set to 8. 

\paragraph{$k$NN}
We use the \texttt{faiss} library \cite{faiss} to run $k$NN on gpu. Currently we are applying the
brute-force exact search method given our problem size is still considered small. The 
\texttt{faiss} library provides optimized $k$-selection routines to speed up the 
computation and for more details we refer the readers to the original article 
\cite{faiss}.

\subsection{Results}
\label{Sec:quantitative_results}
We show quantitative comparison with existing works in Table~\ref{Tab:Results}. 
We have achieved state-of-the-art results on the Sintel clean dataset in the
two-view case, obtaining $11.3\%$ improvement  ($1.94 \rightarrow 1.72$) over RAFT\cite{raft}. 
We also tested the ``warm-start'' strategy in RAFT \cite{raft}, which uses optical flow estimated 
in the previous frames to initialize current optical flow estimation. 
We found that it did not help our performance hence our result is still behind RAFT's 
``warm-start'' results. 
On the Sintel final dataset our result is comparable to state-of-the-art results, currently behind 
RAFT \cite{raft} and DICL \cite{dicl} while better than all the other methods. 
On the KITTI-15 dataset, our result is behind RAFT \cite{raft} and MaskFlowNet \cite{maskflownet}
and supersedes other approaches. We tested the generalization ability of our approach by evaluating
the pre-trained model (C+T) on Sintel and KITTI-15. We have achieved the best results on Sintel 
clean and are second to RAFT \cite{raft} on Sintel final and KITTI-15.

The improvements on Sintel clean can be attributed to the larger correlation volume ($1/4$ 
resolution vs $1/8$ resolution). We provide qualitative results in Figure~\ref{fig:vis}, which
clearly demonstrates the advantage of building the correlation volume and predicting optical 
flow in high resolutions. It can be seen that the motion of fine structures fails to be 
captured by RAFT but can be accurately predicted by our approach, with the use of a $1/4$ resolution
correlation volume. 
With Sintel final and KITTI-15, there exists significantly more motion 
blur and featureless regions. Therefore, setting $k=8$ might be too small to reach the same 
performance as a dense correlation volume. We analyse the effect of $k$ in 
Section~\ref{Sec:Ablations} via ablation experiments. We want to emphasize that even though
we do not outperform RAFT\cite{raft} in all datasets, it is surprising to see that a sparse 
approach can do almost as well given the few storage of correlation values. For each pixel,
we store and process only $k=8$ correlations whereas RAFT requires to store $h \times w$
correlations, limiting its ability to scale up to higher resolutions.

\setlength\tabcolsep{.7em}
\begin{table}[t]
\centering
\resizebox{0.478\textwidth}{!}{
\begin{tabular}{clccccc}
\toprule
 & \multirow{2}{*}{\centering Method} & \multirow{2}{1cm}{\centering Chairs (val)} & \multicolumn{2}{c}{\centering Sintel (train)} & \multicolumn{2}{c}{\centering KITTI-15 (train)} \\
\cmidrule(lr){4-5}
\cmidrule(lr){6-7}
& & & Clean & Final & EPE & F1-all \\
\midrule
\multirow{2}{1.2cm}{\centering Resolution} 
                            & Eighth & 0.95 & 1.55 & 3.07 & 5.74 & 20.2  \\
                            & \underline{Quarter} & 0.71 & 1.29 & 2.95 & 6.80 & 19.3 \\ 
\midrule
\multirow{3}{1.2cm}{\centering Ours Sparsity$^\star$} 
                            & $k=1$ & 1.14  & 1.97 & 3.44 & 7.56 & 25.6 \\
                            & $k=4$ & 0.98 & 1.72 & 3.07 & 6.32 & 21.8 \\ 
                            & \underline{$k=8$} & 0.95 & 1.55 & 3.07 & 5.74 & 20.2 \\ 
\midrule
\multirow{6}{1.2cm}{\centering RAFT Sparsity \cite{raft}}      
                            & $k=1$  & 1.20 & 3.13 & 4.26 & 14.4 & 39.3 \\
                            & $k=8$  & 0.93 & 1.64 & 2.97 & 7.18 & 22.8 \\ 
                            & $k=32$ & 0.87 & 1.50 & 2.82 & 5.77 & 19.5 \\ 
                            & $k=128$ & 0.84 & 1.50 & 2.75 & 5.61 & 19.0 \\ 
                            & ReLU & 0.91 & 1.44 & 2.75 & 5.09 & 16.7 \\ 
                            & Dense & 0.88 & 1.44 & 2.73 & 5.10 & 17.5 \\

\bottomrule
\end{tabular}
}
\caption{\textbf{Ablation experiment results.} \em\small Settings used in our final model are underlined. The details are in  Section~\ref{Sec:Ablations}. We also give results run on RAFT's original code but with varying sparsity levels of the correlation volume. $^\star$We ran these experiments on $1/8$ resolution. }
\label{Tab:Ablations}
\end{table}
\setlength\tabcolsep{.7em}
\begin{table}[ht]
\begin{subtable}[ht]{0.47\textwidth}
\centering
\resizebox{0.85\textwidth}{!}{
\begin{tabular}{lcccc}
\toprule
 \multirow{2}{*}{\centering Sparsity} & \multicolumn{2}{c}{\centering $1/4$ Resolution} & \multicolumn{2}{c}{\centering $1/8$ Resolution} \\
\cmidrule(lr){2-3}
\cmidrule(lr){4-5}
& Size & Memory & Size & Memory \\
\midrule
Dense & $7.8 \times 10^8$ & $\SI{3.1}{\giga\byte}$ & $4.8 \times 10^7$ & $\SI{191}{\mega\byte}$ \\
$k=8$ & $2.2 \times 10^5$ & $\SI{0.9}{\mega\byte}$ & $5.5 \times 10^4$ & $\SI{0.2}{\mega\byte}$ \\ 
$k=32$ & $8.9 \times 10^5$ & $\SI{3.6}{\mega\byte}$ & $2.2 \times 10^5$ & $\SI{0.9}{\mega\byte}$ \\
$k=128$ & $3.6 \times 10^6$ & $\SI{14.3}{\mega\byte}$ & $8.8 \times 10^5$ & $\SI{3.5}{\mega\byte}$ \\ 
\bottomrule
\end{tabular}
}
\caption{\em \small Size and memory of a correlation volume based on a pair of images of size $436 \times 1024$. Size refers to the number of elements and the correlation volumes are stored in 32-bit floats.}
\vspace{0.3cm}
\label{Tab:Theoretical_memory}
\end{subtable}
\hfill
\begin{subtable}[ht]{0.47\textwidth}
    \centering
    \resizebox{0.55\textwidth}{!}{
    \begin{tabular}{ccc}
    \toprule 
        Method & batch = 1 & batch = 2  \\
        \midrule
        RAFT \cite{raft} & 10.6 GB & 20.0 GB \\
        Ours & 6.1 GB & 9.3 GB \\
    \bottomrule
    \end{tabular}
    }
    \caption{\em \small Actual memory consumption when training on the Sintel dataset. The
    correlation volumes are built from $1/4$ resolution feature maps. We use a random crop
    of $400 \times 720$. The batch size is set to $1$ and $2$. }
    \vspace{0.3cm}
    \label{Tab:Actual_memory}
\end{subtable}

\caption{\textbf{Results for memory consumption.}}
\label{Tab:Memory}
\end{table}
\subsection{Ablation Study}
\label{Sec:Ablations}
We conducted ablation experiments to validate our hypothesis that top-$k$ correlations
are sufficient to give a good representation of the full correlation volume. The main
setting here is how large should $k$ be. We show results on our model with
different choices of $k$ on $1/8$ resolutions. It can be clearly seen that larger $k$ gives
better performance. Even when $k = 1$, the results are still reasonable and do not completely fail. 

We also compare $1/4$ resolution correlation volume with $1/8$ resolution correlation
volume and we can see that $1/4$ resolution correlation volume gives better results in
all datasets except the EPE in KITTI-15. Since large correlation volumes are constructed from
higher-resolution feature maps, we believe that larger correlation volumes are more descriptive
of the image details and the results agree with our hypothesis. 

An additional experiment we conducted is with RAFT's original implementation. We keep the top-$k$
elements in the correlation volume and set the rest to be zero. We vary $k$ to be $\{1, 8, 32, 128\}$.
In Table \ref{Tab:Ablations}, ReLU refers to setting all negative values to be zero and only keeping the positive correlations. We 
also trained with the original code which is denoted as ``Dense''. We can see that larger $k$ gives
better results and $k=\{32, 128\}$ almost reach the same performance as the dense method. ReLU even
outperforms the dense method on KITT-15. This again validates our hypothesis that there exists significant
redundancy in the current dense approach and a sparse correlation volume with a large enough $k$ could do
just as well. 

\subsection{Memory Consumption}
Our method of processing the sparse correlation volume does not introduce new 
learning parameters. 
The number of parameters in our network is $\SI{5.3}{\mega\byte}$, which is the same 
as RAFT. Given an image pair of size $436 \times 1024$, the size and memory of
sparse correlation volumes and dense correlation volumes in $1/4$ and $1/8$ 
resolutions are listed in Table~\ref{Tab:Theoretical_memory}. 

When correlation volumes are built from $1/8$ resolution feature maps, our approach does
not lead to a significant memory saving. This is due to the constant $\SI{2}{\giga\byte}$ memory 
overhead of the $k$NN search library and also the correlation volume is not a memory
bottleneck ($\SI{191}{\mega\byte}$ when batch size = 1) when resolutions are small.

However, our approach demonstrates a clear advantage when correlation volumes are built from $1/4$
resolution feature maps for images of size $436 \times 1024$. When training
at $1/4$ resolution, with a random crop of $400 \times 720$ of the original image 
and batch size = 1 and 2,
our approach consumes around $50\%$ of total memory compared to RAFT. The results are demonstrated
in Table~\ref{Tab:Actual_memory}. This showcases the effectiveness of our approach in
saving memory when correlation volumes are scaled to higher resolutions. 

\subsection{Limitations}
\begin{figure*}[h!]
     \centering
     \begin{subfigure}[b]{0.195\textwidth}
         \centering
         \includegraphics[width=\textwidth]{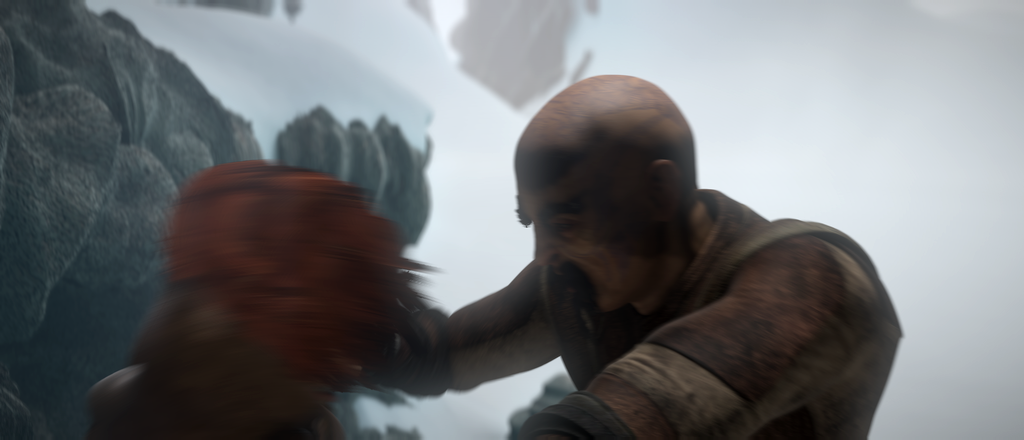}
         \caption{Image 1}
     \end{subfigure}
     \hfill
     \begin{subfigure}[b]{0.195\textwidth}
         \centering
         \includegraphics[width=\textwidth]{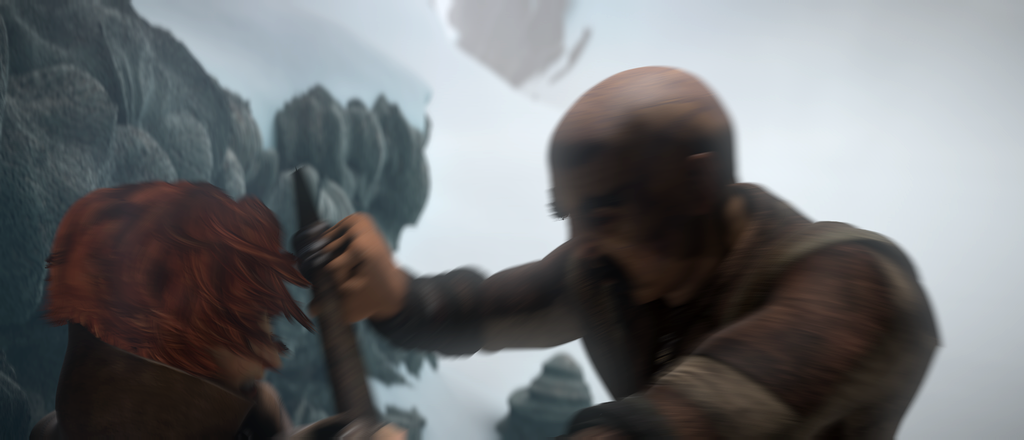}
         \caption{Image 2}
     \end{subfigure}
     \hfill
     \begin{subfigure}[b]{0.195\textwidth}
         \centering
         \includegraphics[width=\textwidth]{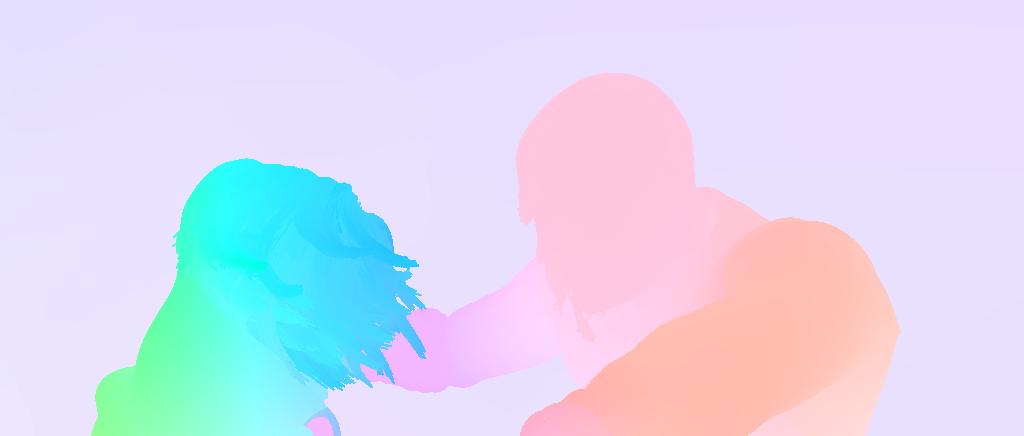}
         \caption{GT}
     \end{subfigure}
     \hfill
     \begin{subfigure}[b]{0.195\textwidth}
         \centering
         \includegraphics[width=\textwidth]{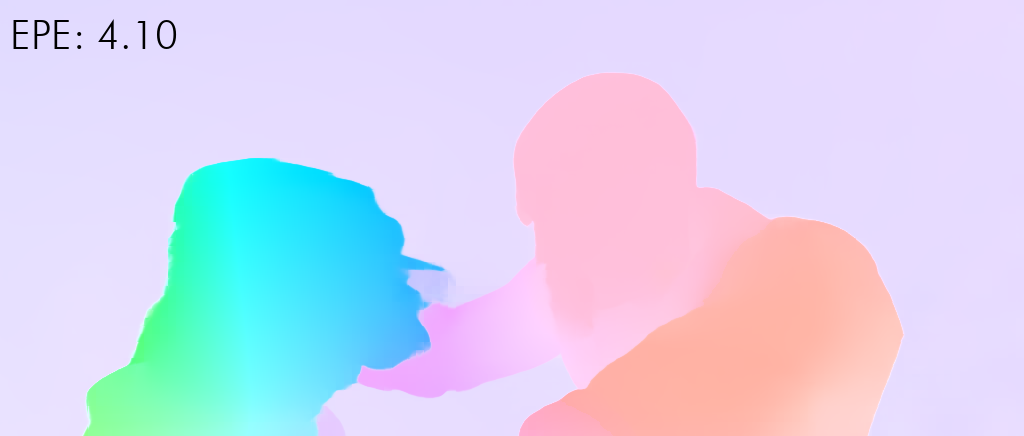}
         \caption{RAFT}
     \end{subfigure}
     \hfill
     \begin{subfigure}[b]{0.195\textwidth}
         \centering
         \includegraphics[width=\textwidth]{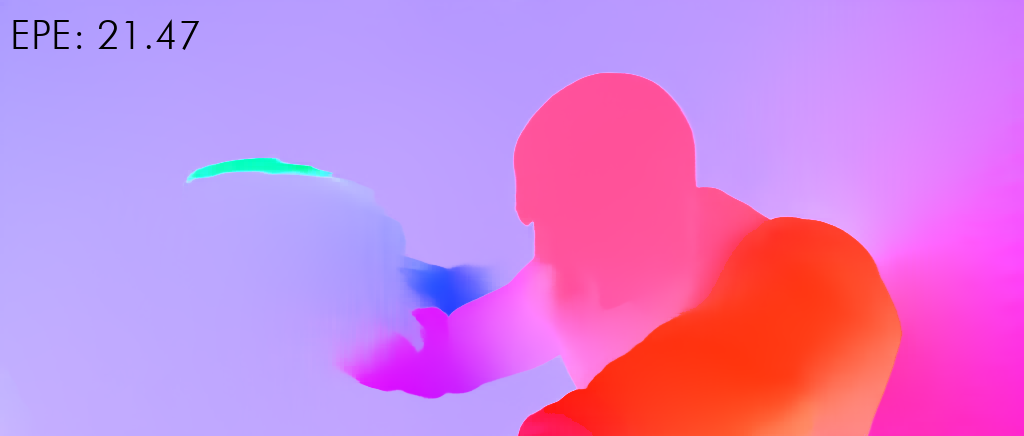} 
         \caption{Ours}
     \end{subfigure}
     \caption{\textbf{An example failure case.} \em \small When the scene contains significant
     featureless regions or motion blur, top-$k$ correlations may not contain the correct matches
     and can lead to incorrect flow prediction. }
     \label{fig:failure_case}
\end{figure*}

Increasing the resolution to $1/4$, we have observed consistent 
improvements on fine-structure motions (\eg the bamboo sequence in Sintel). 
However, the commonly used metric for overall evaluation, mean EPE, is defined to be biased 
towards large 
motions on large regions. One particular weakness of our approach is the 
handling of featureless or blurry regions. 
Such features typically have a large number of matches due to the ambiguity, top-$k$ might not be 
sufficient to cover the correct match
and could give misguided motion prediction. An example failure case is shown in 
Figure~\ref{fig:failure_case}.
One can see that the red hair contains significant motion
blur, where our top-$k$ correlations do not contain the correct matches hence lead to
incorrect prediction.

\section{Related Work}
\label{Sec:Related}

Optical flow was first formulated as a continuous optimization problem with
variational approaches \cite{horn}. Various subsequent papers have worked on
on improving robustness \cite{black} and energy term \cite{tvl1}, incorporating 
descriptor matching into energy minimization \cite{brox} and improving
regularizations \cite{ranftl}. Accurate flow fields can be predicted when
displacements are small. However, their performances are limited at large
displacements, due to the use of first order Taylor approximation.

Pyramidal approaches were developed to handle large displacements in
stereo and optical flow, pioneered by Quam \etal \cite{quam}. Traditional methods
build a Gaussian pyramid \cite{burt1983laplacian} and predict optical flow or stereo
in a coarse-to-fine manner. In contrast, deep optical flow approaches
\cite{pwcnet,hd3,liteflownet,liteflownet2,irr,vcn,scopeflow,maskflownet,dicl}
build a feature pyramid to extract more representative information on 
different levels through learning. Optical flows are then predicted in a
coarse-to-fine manner. A local correlation volume with a limited search range is
constructed in each level, based on featured maps warped with the upsampled flow
from the previous level. This approach limits the search range of the correlation
volume and effectively reduces the memory and computational cost for processing it.
However, as pointed out in previous
works \cite{devon, improve, raft}, pyramidal warping approaches can have ghosting 
effects at occlusions. They are also limited at handling the small object large
motion problem and fine-level predictions often fail to recover errors made in 
coarser levels. By contrast, our method operates at a single resolution and does
not suffer from such problems. 

Pyramidal approaches are designed to handle large motions while keeping memory
cost small. However, they are not the only successful approach to dealing with large
motions.
Before the deep learning era, optical flow estimation was formulated as a
discrete optimization problem by solving a Markov random field (MRF) \cite{boykov}.
Chen \etal \cite{fullflow} proposed a one-shot global discrete optimization approach
at a single resolution with a distance transform, which is then refined 
using continuous optimization. 
Menze \etal \cite{discreteflow} proposed reducing the search
space by limiting to a fixed number $k$ matches per pixel via approximate
nearest neighbour search. Our idea is similar in the sense that we both limit
the search space to top-$k$, but rather than solving an MRF, we propose
explicitly constructing a sparse correlation volume and using iterative refinement
to predict optical flow. The main difference is that our final solution does not
necessarily lie the top-$k$ solution space, which gives better occlusion handling.
Our idea is also inspired by the recent paper on learning to find sparse matches,
which proposed to find top-$k$ matches first then process with sparse convolutions. 
However, we do not use sparse convolutions but rather convert the sparse correlation
volume to a dense tensor with our proposed multi-scale displacement encoder. 

A recent breakthrough in deep optical flow estimation has been achieved by RAFT
\cite{raft}, which proposed constructing a dense all-pairs correlation volume 
on a single resolution and adopting a recurrent network to iteratively predict 
optical flow. Due to the memory cost of the correlation volume, the feature map
resolutions are limit to $1/8$ of the original image resolution. Our approach is
different to RAFT in three major ways:
\begin{enumerate}
    \item Rather than constructing a dense all-pairs correlation volume, we
    construct a sparse correlation volume where only the top $k$ correlations are
    stored. This allows us to reduce the spatial complexity from $O(N^2)$ to $O(N)$, where $N$ refers to the number of pixels of an image.
    \item Because of the savings in memory cost, we can build the correlation volume
    from higher resolution feature maps ($1/4$ vs. $1/8$) without limiting the 
    searching range. This allows our method to accurately predict the motion
    of the finer structures. 
    \item We propose a new way of iteratively decoding the sparse correlation
    volume. Rather than sampling in a dense correlation volume at different 
    locations, we iteratively update the coordinates of the sparse correlation 
    volume and apply ``bilinear splatting'' to splat the correlations onto the 
    integer grids. 
\end{enumerate}

\section{Conclusions}
Since the publication of RAFT \cite{raft}, the use of all-pairs (large
displacements) correlation volume is becoming a standard way of solving for
optical flow due to its superior performance compared to pyramidal approaches.
However, the memory consumption of an all-pairs correlation volume grows 
quadratically with the number of pixels, quickly limiting its ability to handle 
high-resolution images or capture fine-structure motion. 
We observed a rather surprising fact where storing just the top-$k$ correlations
provides almost as good results as storing the dense correlation volume. 
The sparse correlation volume method proposed in this paper provides an alternative
approach to store the all-pairs matching information which massively reduces 
memory consumption while gives accurate prediction of optical flow. 
Experiments validated the feasibility of using sparse correlation volume in optical
flow estimation tasks. 
We believe our paper has paved a way for future optical flow research directions,
where the memory requirement of correlation volumes is no longer a limiting factor.

\section*{Acknowledgements}
This research is funded in part by the ARC Centre of Excellence for Robotic Vision 
(CE140100016), ARC Discovery Project grant (DP200102274) and (DP190102261). We thank
all reviewers for their valuable comments. 

{\small
\bibliographystyle{ieee_fullname}
\bibliography{egbib}

\begin{thebibliography}{10}\itemsep=-1pt

\bibitem{scopeflow}
Aviram Bar-Haim and Lior Wolf.
\newblock Scopeflow: Dynamic scene scoping for optical flow.
\newblock {\em CVPR}, 2020.

\bibitem{black}
Michael~J Black and Padmanabhan Anandan.
\newblock A framework for the robust estimation of optical flow.
\newblock {\em ICCV}, 1993.

\bibitem{boykov}
Yuri Boykov and Vladimir Kolmogorov.
\newblock An experimental comparison of min-cut/max-flow algorithms for energy
  minimization in vision.
\newblock {\em TPAMI}, 2004.

\bibitem{brox}
Thomas Brox, Christoph Bregler, and Jitendra Malik.
\newblock Large displacement optical flow.
\newblock {\em CVPR}, 2009.

\bibitem{burt1983laplacian}
Peter Burt and Edward Adelson.
\newblock The laplacian pyramid as a compact image code.
\newblock {\em IEEE Transactions on communications}, 1983.

\bibitem{sintel}
D.~J. Butler, J. Wulff, G.~B. Stanley, and M.~J. Black.
\newblock A naturalistic open source movie for optical flow evaluation.
\newblock {\em ECCV}, 2012.

\bibitem{fullflow}
Qifeng Chen and Vladlen Koltun.
\newblock Full flow: Optical flow estimation by global optimization over
  regular grids.
\newblock {\em CVPR}, 2016.

\bibitem{minkowski}
Christopher Choy, JunYoung Gwak, and Silvio Savarese.
\newblock 4d spatio-temporal convnets: Minkowski convolutional neural networks.
\newblock {\em CVPR}, 2019.

\bibitem{flownet}
Alexey Dosovitskiy, Philipp Fischer, Eddy Ilg, Philip Hausser, Caner Hazirbas,
  Vladimir Golkov, Patrick Van Der~Smagt, Daniel Cremers, and Thomas Brox.
\newblock Flownet: Learning optical flow with convolutional networks.
\newblock {\em ICCV}, 2015.

\bibitem{improve}
Markus Hofinger, Samuel~Rota Bul{\`o}, Lorenzo Porzi, Arno Knapitsch, Thomas
  Pock, and Peter Kontschieder.
\newblock Improving optical flow on a pyramid level.
\newblock {\em ECCV}, 2020.

\bibitem{horn}
Berthold~KP Horn and Brian~G Schunck.
\newblock Determining optical flow.
\newblock {\em Techniques and Applications of Image Understanding}, 1981.

\bibitem{hosni2012fast}
Asmaa Hosni, Christoph Rhemann, Michael Bleyer, Carsten Rother, and Margrit
  Gelautz.
\newblock Fast cost-volume filtering for visual correspondence and beyond.
\newblock {\em TPAMI}, 2012.

\bibitem{liteflownet}
Tak-Wai Hui, Xiaoou Tang, and Chen Change~Loy.
\newblock Liteflownet: A lightweight convolutional neural network for optical
  flow estimation.
\newblock {\em CVPR}, 2018.

\bibitem{liteflownet2}
Tak-Wai Hui, Xiaoou Tang, and Chen~Change Loy.
\newblock A lightweight optical flow cnn-revisiting data fidelity and
  regularization.
\newblock {\em TPAMI}, 2020.

\bibitem{irr}
Junhwa Hur and Stefan Roth.
\newblock Iterative residual refinement for joint optical flow and occlusion
  estimation.
\newblock {\em CVPR}, 2019.

\bibitem{flownet2}
Eddy Ilg, Nikolaus Mayer, Tonmoy Saikia, Margret Keuper, Alexey Dosovitskiy,
  and Thomas Brox.
\newblock Flownet 2.0: Evolution of optical flow estimation with deep networks.
\newblock {\em CVPR}, 2017.

\bibitem{faiss}
Jeff Johnson, Matthijs Douze, and Herv{\'e} J{\'e}gou.
\newblock Billion-scale similarity search with gpus.
\newblock {\em arXiv preprint arXiv:1702.08734}, 2017.

\bibitem{gcnet}
Alex Kendall, Hayk Martirosyan, Saumitro Dasgupta, Peter Henry, Ryan Kennedy,
  Abraham Bachrach, and Adam Bry.
\newblock End-to-end learning of geometry and context for deep stereo
  regression.
\newblock {\em ICCV}, 2017.

\bibitem{hd1k}
Daniel Kondermann, Rahul Nair, Katrin Honauer, Karsten Krispin, Jonas Andrulis,
  Alexander Brock, Burkhard Gussefeld, Mohsen Rahimimoghaddam, Sabine Hofmann,
  Claus Brenner, et~al.
\newblock The hci benchmark suite: Stereo and flow ground truth with
  uncertainties for urban autonomous driving.
\newblock {\em CVPR Workshop}, 2016.

\bibitem{devon}
Yao Lu, Jack Valmadre, Heng Wang, Juho Kannala, Mehrtash Harandi, and Philip
  Torr.
\newblock Devon: Deformable volume network for learning optical flow.
\newblock {\em WACV}, 2020.

\bibitem{things}
N. Mayer, E. Ilg, P. H{\"a}usser, P. Fischer, D. Cremers, A. Dosovitskiy, and
  T. Brox.
\newblock A large dataset to train convolutional networks for disparity,
  optical flow, and scene flow estimation.
\newblock {\em CVPR}, 2016.

\bibitem{discreteflow}
Moritz Menze, Christian Heipke, and Andreas Geiger.
\newblock Discrete optimization for optical flow.
\newblock {\em GCPR}, 2015.

\bibitem{kitti}
Moritz Menze, Christian Heipke, and Andreas Geiger.
\newblock Joint 3d estimation of vehicles and scene flow.
\newblock {\em ISPRS Workshop on Image Sequence Analysis (ISA)}, 2015.

\bibitem{pytorch}
Adam Paszke, Sam Gross, Soumith Chintala, Gregory Chanan, Edward Yang, Zachary
  DeVito, Zeming Lin, Alban Desmaison, Luca Antiga, and Adam Lerer.
\newblock Automatic differentiation in pytorch.
\newblock {\em NIPS Workshop}, 2017.

\bibitem{quam}
Lynn~H Quam.
\newblock Hierarchical warp stereo.
\newblock {\em Readings in Computer Vision}, 1987.

\bibitem{ranftl}
Ren{\'e} Ranftl, Kristian Bredies, and Thomas Pock.
\newblock Non-local total generalized variation for optical flow estimation.
\newblock {\em ECCV}, 2014.

\bibitem{pwcnet}
Deqing Sun, Xiaodong Yang, Ming-Yu Liu, and Jan Kautz.
\newblock Pwc-net: Cnns for optical flow using pyramid, warping, and cost
  volume.
\newblock {\em CVPR}, 2018.

\bibitem{pwcnet+}
Deqing Sun, Xiaodong Yang, Ming-Yu Liu, and Jan Kautz.
\newblock Models matter, so does training: An empirical study of cnns for
  optical flow estimation.
\newblock {\em TPAMI}, 2019.

\bibitem{raft}
Zachary Teed and Jia Deng.
\newblock Raft: Recurrent all-pairs field transforms for optical flow.
\newblock {\em ECCV}, 2020.

\bibitem{dicl}
Jianyuan Wang, Yiran Zhong, Yuchao Dai, Kaihao Zhang, Pan Ji, and Hongdong Li.
\newblock Displacement-invariant matching cost learning for accurate optical
  flow estimation.
\newblock {\em NeurIPS}, 2020.

\bibitem{dcflow}
Jia Xu, Ren{\'e} Ranftl, and Vladlen Koltun.
\newblock Accurate optical flow via direct cost volume processing.
\newblock {\em CVPR}, 2017.

\bibitem{xu2011motion}
Li Xu, Jiaya Jia, and Yasuyuki Matsushita.
\newblock Motion detail preserving optical flow estimation.
\newblock {\em TPAMI}, 2011.

\bibitem{vcn}
Gengshan Yang and Deva Ramanan.
\newblock Volumetric correspondence networks for optical flow.
\newblock {\em NeurIPS}, 2019.

\bibitem{hd3}
Zhichao Yin, Trevor Darrell, and Fisher Yu.
\newblock Hierarchical discrete distribution decomposition for match density
  estimation.
\newblock {\em CVPR}, 2019.

\bibitem{tvl1}
Christopher Zach, Thomas Pock, and Horst Bischof.
\newblock A duality based approach for realtime tv-l 1 optical flow.
\newblock {\em Joint Pattern Recognition Symposium}, 2007.

\bibitem{maskflownet}
Shengyu Zhao, Yilun Sheng, Yue Dong, Eric~I Chang, Yan Xu, et~al.
\newblock Maskflownet: Asymmetric feature matching with learnable occlusion
  mask.
\newblock {\em CVPR}, 2020.

\end{thebibliography}
}

\end{document}